\documentclass[12pt]{iopart}
\usepackage{cite}
\usepackage[linesnumbered,ruled]{algorithm2e}\SetKwRepeat{Do}{do}{while}%
\usepackage{graphicx}
\expandafter\let\csname equation*\endcsname\relax
\expandafter\let\csname endequation*\endcsname\relax
\usepackage{amsmath}
\usepackage{amsfonts} 
\usepackage{subfigure}
\usepackage{tikz} 
\DeclareMathOperator*{\argminA}{arg\,min} 

\begin{document}
\title[Measurement Science and Technology]{Real-Time Stereo Vision-Based Lane Detection System}

\author{Rui Fan \& Naim Dahnoun}

\address{Department of Electrical and Electronic Engineering, Merchant Venturers Building, University of Bristol, Bristol, BS8 1UB, UK}
\ead{\{ranger.fan, naim.dahnoun\}@bristol.ac.uk}
\vspace{10pt}
\begin{indented}
\item[]February 2018
\end{indented}

\begin{abstract}
The detection of multiple curved lane markings on a non-flat road surface is still a challenging task for automotive applications. To make an improvement, the depth information can be used to greatly enhance the robustness of the lane detection systems. The proposed system in this paper is developed from our previous work where the dense vanishing point $\boldsymbol{V_p}$ is estimated globally to assist the detection of multiple curved lane markings. However, the outliers in the optimal solution may severely affect the accuracy of the least squares fitting when estimating $\boldsymbol{{V}_p}$. Therefore, in this paper we use Random Sample Consensus to update the inliers and outliers iteratively until the fraction of the number of inliers versus the total number exceeds our pre-set threshold. This significantly helps the system to overcome some suddenly changing conditions. Furthermore, we propose a novel lane position validation approach which provides a piecewise weight $w_g$ based on $\boldsymbol{{V}_p}$ and the gradient $\nabla$ to reduce the gradient magnitude of the non-lane candidates. Then, we compute the energy of each possible solution and select all satisfying lane positions for visualisation. The proposed system is implemented on a heterogeneous system which consists of a Intel Core i7-4720HQ CPU and a NVIDIA GTX 970M GPU. A processing speed of 143 fps has been achieved, which is over 38 times faster than our previous work. Also, in order to evaluate the detection precision, we tested 2495 frames with 5361 lanes from the KITTI database (1637 lanes more than our previous experiment). It is shown that the overall successful detection rate is improved from $98.7\%$ to $99.5\%$.


\end{abstract}

\section{Introduction}
\label{sec.ld_introduction}

Various prototype vehicle road tests have been conducted by Google in the US since 2012, and its subsidiary X plans to commercialise their autonomous cars as from 2020 \cite{Rosenzweig2015}. Recently, Volvo has been planning to conduct a series of self-driving experiments involving about 100 cars in China and many companies like Ford and Uber have entered the race to make driver-less taxis a reality. The techniques like lane detection systems in the ADAS (Advanced Driver Assistance Systems) are playing an increasingly crucial role in enhancing driving safety and minimising the possibilities of fatalities.

Current lane detection algorithms can mainly be grouped into two categories: feature-based and model-based \cite{Narote2018}. The feature-based algorithms extract the local, meaningful and detectable parts of the image, such as edges, texture and colour, to segment lanes and road boundaries from the background pixel by pixel \cite{Bertozzi1998}. On the other hand, the model-based algorithms try to represent the lanes with a mathematical equation based upon some assumptions of the road geometry \cite{Wang2004}. The commonly used lane models include: linear, parabolic, linear-parabolic and splines. The linear model works well for the lanes with a low curvature, as demonstrated by the linear lane detection system in our previous work \cite{Fan2016, Ozgunalp2014}. However, a more flexible road model is inevitable when the lanes with a higher curvature exist. Therefore, some algorithms \cite{Kluge1995, Wang2008, Zhou2006, Kreucher1999} use a parabolic model to represent the lanes with a constant curvature. For some other more complicated cases, Jung et al. propose a linear-parabolic combined lane model, where the nearby lanes are represented as linear models, whereas the far ones are modelled as parabolas \cite{Jung2005}. In addition to the models mentioned above, the spline model is another alternative which interpolates the pixels on a lane as an arbitrary shape \cite{Wang2004, Wang2000}. However, the more parameters introduced into a flexible model, the higher will be the  computational complexities of the algorithm. Therefore, we turn our focus on some additional important properties of 3D imaging techniques instead of being limited to 2D information. 

One of the most prevalently used methods is IPM (Inverse Perspective Mapping). With the assumption that two lanes are parallel to each other in the world coordinate system (WCS), IPM is able to map a 3D scenery into a 2D bird's eye view \cite{Nieto2007}. Furthermore, many researchers \cite{Schreiber2005, Hanwell2012, Fardi2004, Wang2012} propose to use the vanishing point $\boldsymbol{{V}_p}=[V_{px}, V_{py}]^\top$ to model lane markings and road boundaries. However, their algorithms work well only if we assume the road surface is flat or the camera parameters are known. Therefore, we pay closer attention to the disparity information provided by either active sensors (radar and laser) or passive sensors (stereo cameras) \cite{Ozgunalp2017}. Since Labayrade et al. proposed the concept of "v-disparity" \cite{Labayrade2002}, disparity information has been widely used to enhance the robustness of the lane detection systems. Our previous work \cite{Ozgunalp2017} shows a particular instance where the disparity information is successfully combined with a lane detection algorithm to estimate $V_{py}$ for a non-flat road surface. At the same time, a lot of redundant information can be eliminated from the obstacles by comparing their disparities with the theoretical ones. However, the estimation of $V_{py}$ suffers from the outliers in an optimal solution, and the lanes are sometimes unsuccessfully detected because the selection of plus-minus peaks is not always effective. Moreover, real-time performance is still a challenging task in \cite{Ozgunalp2017} because of the intensive computational complexities of the algorithm. Therefore, in this paper we present an improved lane detection system for the problems mentioned above. 

\begin{figure}[!t]
	\centering
	\includegraphics[width=1\textwidth]{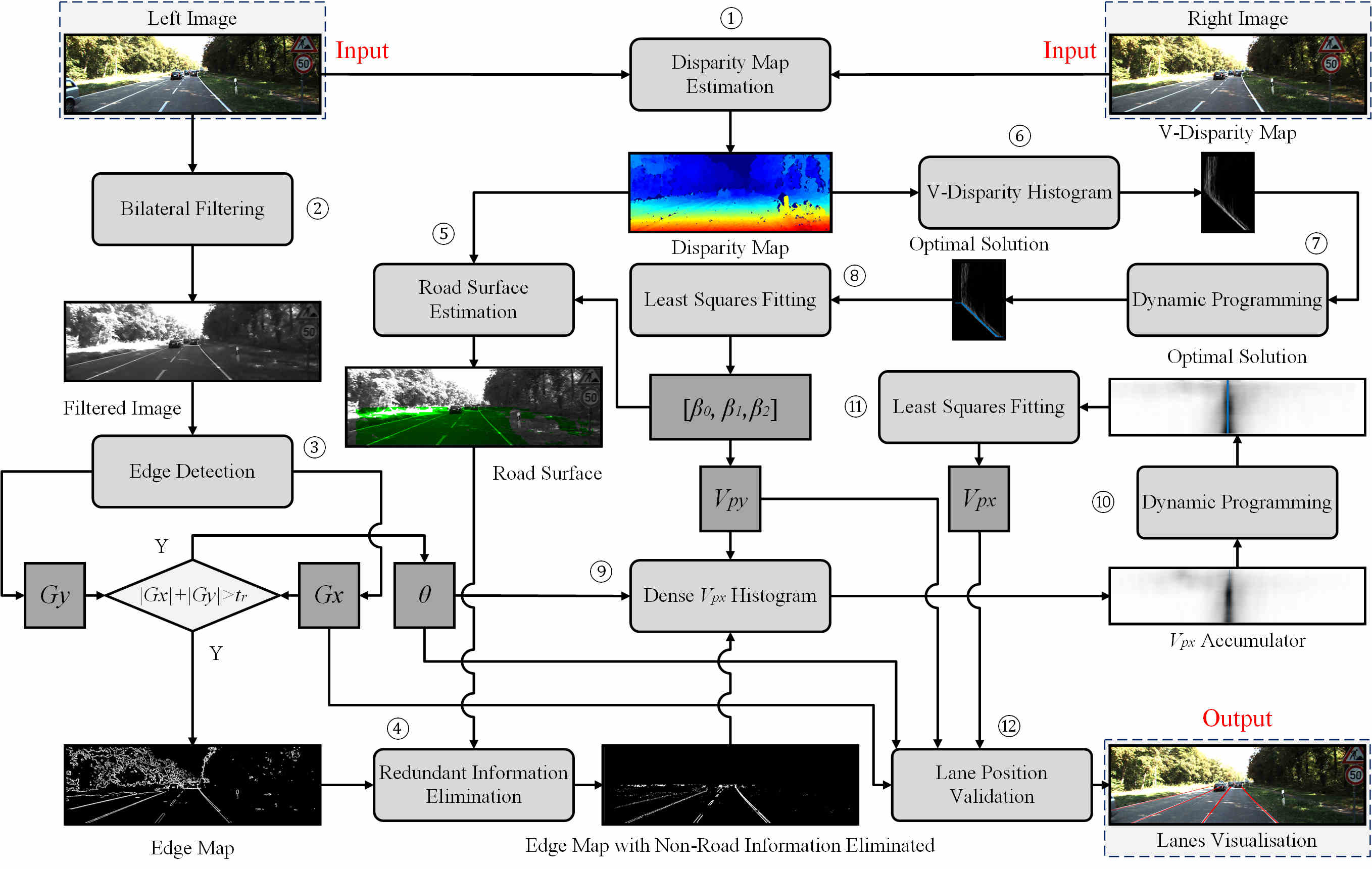}
	\caption{The block diagram of the proposed lane detection system.}
	\label{fig.ld_block_diagram}
\end{figure}

The proposed multiple lane detection system is composed of four main components: disparity map estimation, dense $V_{py}$ estimation, dense $V_{px}$ estimation and lane position validation. The block diagram of the proposed system is illustrated in Fig. \ref{fig.ld_block_diagram}. Procedures 1 to 6 run on a GPU because they are more efficient for parallel processing. On the contrary, the serially-efficient procedures 7 to 12 execute on a CPU . 

Firstly, we estimate the disparity map from a pair of well-calibrated stereo images. The disparity map is mainly used for:
\begin{itemize}
	\item estimation of dense $V_{py}$,
	\item road surface detection, and
	\item elimination of the noise from obstacles.
\end{itemize}

The disparity information is combined with the lane detection system by analysing a so-called v-disparity histogram. In this paper, we assume the road surface is non-flat and the projection of on-road disparities on the v-disparity map can be represented by a parabola $f(v)=\beta_0+\beta_1v+\beta_2v^2$, where $\boldsymbol{\beta}=[\beta_0,\beta_1,\beta_2]^\top$ is the parameters of the vertical road profile and $v$ is the row number. Compared with various quadratic pattern detectors, dynamic programming (DP) provides a more efficient way of extracting the best path from the v-disparity map by minimising the energy of an optimisation problem. Then, the extracted path $\boldsymbol{M_{y}}$ is fitted to a parabola using the least squares fitting (LSF). 
 However, the outliers $\mathcal{O}$ in $\boldsymbol{M_{y}}$ may affect the accuracy of the LSF significantly. Therefore, we propose to update the parabola function iteratively using RANSAC (Random Sample Consensus) until the capacity of inliers $\mathcal{I}$ does not change, which greatly helps the system maintain  the robustness when estimating $V_{py}$. Furthermore, we employ a bilateral filter before approximating the horizontal and vertical derivatives: $G_x$ and $G_y$. 
 This achieves a better performance than the median filter which was used in our previous work \cite{Ozgunalp2017} in terms of the edge-preservation and noise elimination. $V_{py}$ and the orientation of each pixel in the road surface area are then used to estimate $V_{px}$, where we apply the same strategy of RANSAC to minimise the influence of outliers $\mathcal{O}$ on the LSF. An arbitrary lane marking or road boundary can thus be extracted using the information of $\boldsymbol{V_p}=[V_{px}, V_{py}]^\top$. To locate the correct lane markings, we propose a novel lane position validation approach using the information of $G_x$ (the dark-light transition of the lane markings has a positive and higher $G_x$ than the non-edge pixels, while $-G_x$ is positive and higher on the light-dark transition of the lane markings). The angle $\angle(\boldsymbol{m},\boldsymbol{n})$ between the orientation vector $\boldsymbol{m}$, obtained from the edge detector and the orientation vector $\boldsymbol{n}$, obtained from $\boldsymbol{V_p}$, is used to provide a piecewise weight $w_g$ to update $G_x$, where the value of an edge pixel that does not belong to either lane markings or road boundaries is cut down. After computing the energy of each possible solution, the satisfying local minima are validated as lane positions. The proposed system achieves a real-time execution of $143$ fps along with an approximately $99.5\%$ successful detection rate on our heterogeneous system consisting of an Intel Core i7-4720HQ CPU and a NVIDIA GTX 970M GPU.

The remainder of the paper is structured as follows: section \ref{sec.ld_system_description} describes the proposed lane detection system, section \ref{sec.ld_experimental_results} evaluates the experimental results, and section \ref{sec.ld_conclusion} concludes the paper.

\section{System Description}
\label{sec.ld_system_description}
\subsection{Disparity map estimation}
\label{sec.disparity_map_estimation}

As compared to many other stereo matching algorithms which are aimed at automotive applications, the trade-off between accuracy and speed has been greatly improved in our previous work \cite{Zhang2013, Fan2017}. Therefore, we employ the algorithm in \cite{Fan2017} to acquire the disparity information for the proposed lane detection system. 

\subsubsection{Memorisation}
\label{sec.block_matching_memorisation}

Due to the insensitivity to the intensity difference, the Normalised Cross-Correlation (NCC) is utilised as the cost function to measure the similarity between two blocks, as shown in Eq. \ref{eq.ncc}. Each block chosen from the left image is matched with a series of blocks on the same epipolar line in the right image \cite{Dahnoun}. The block pair with the highest correlation cost is selected as the best correspondence, and the shifting distance between them is defined as the disparity $d$. 

\begin{equation}
c(u,v,d)=\frac{1}{n\sigma_{l}\sigma_{r}}{\sum\limits_{i=u-\rho}^{i=u+\rho}\sum\limits_{j=v-\rho}^{j=v+\rho}(I_{l}(i,j)-{\mu_{l}})(I_{r}(i-d,j)-{\mu_{r}})}
\label{eq.ncc}
\end{equation}
where $c$ is defined as the correlation cost, and $I_l$ and $I_r$ represent the pixel intensities in the left and right images, respectively. The centre of the block is $(u,v)$ and the side length of the block is $2\rho+1$. $n=(2\rho+1)^2$ represents the number of pixels within each block. ${\mu_{l}}$ and ${\mu_{r}}$ denote the means of the intensities within the left and right blocks, respectively. $\sigma_{l}$ and $\sigma_{r}$ represent their corresponding standard deviations \cite{Fan2017}:


\begin{equation}
\sigma_{l}=\sqrt{\sum\limits_{i=u-\rho}^{i=u+\rho}\sum\limits_{j=v-\rho}^{j=v+\rho}(I_{l}(i,j)-{\mu_{l}})^2/n}
\label{eq.sigma_l}
\end{equation}
\begin{equation}
\sigma_{r}=\sqrt{\sum\limits_{i=u-\rho}^{i=u+\rho}\sum\limits_{j=v-\rho}^{j=v+\rho}(I_{r}(i-d,j)-{\mu_{r}})^2/n}
\label{eq.sigma_r}
\end{equation}

\begin{figure}[t!]
	\centering
	\subfigure[]
	{
		\includegraphics[width=0.46\textwidth]{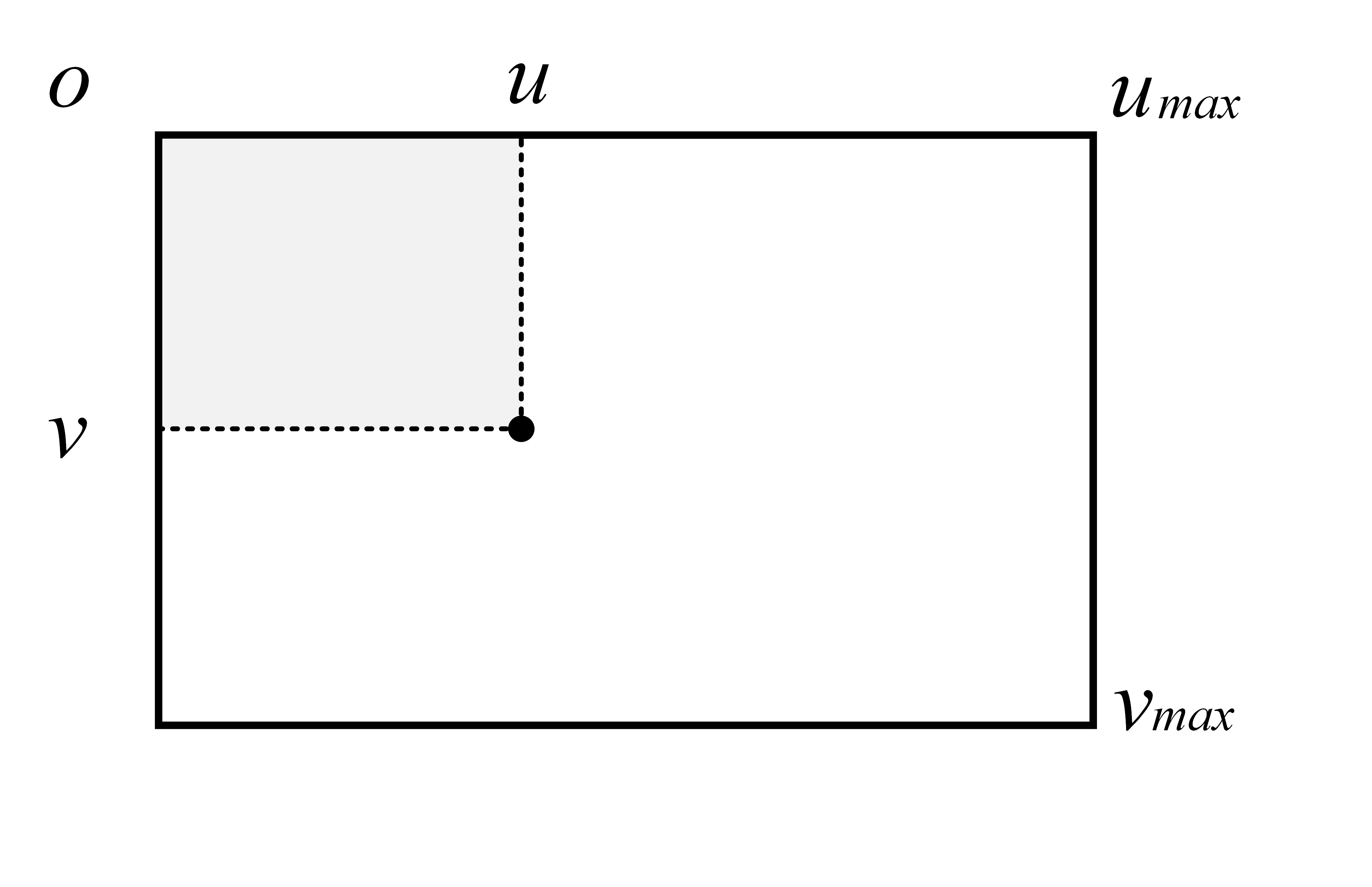}
		\label{fig.integral_image1}
	}	
	\subfigure[]
	{
		\includegraphics[width=0.46\textwidth]{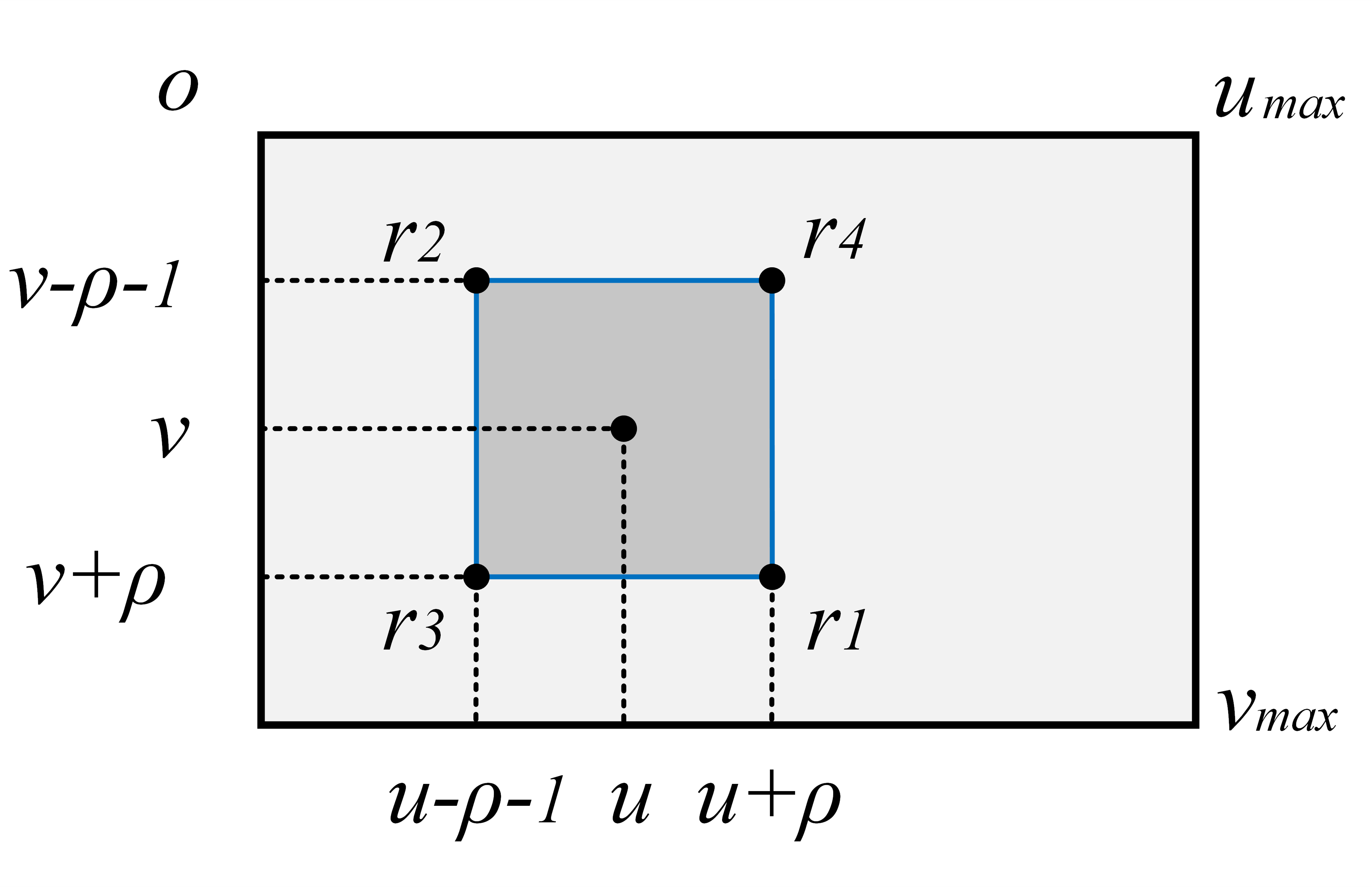}
		\label{fig.integral_image2}
	}

	\caption{Integral image processing. (a) original image. (b) integral image. }
	\label{fig.integral_image}
\end{figure}

When the left block is selected, the calculations of $\mu_l$ and $\sigma_l$ are always repeated because $d$ is only used to select the position of the right blocks for stereo matching. Therefore, the four independent parts $\mu_l$, $\mu_r$, $\sigma_l$ and $\sigma_r$ can be pre-calculated and stored in static program storage for direct indexing. The integral image algorithm can be used to compute $\mu_l$ and $\mu_r$ efficiently \cite{Lewis1995}, which is illustrated in Fig. \ref{fig.integral_image}. The algorithm has two steps: integral image initialisation and value indexing from the initialised reference. In the first step, for a discrete image $I$ whose pixel intensity at $(u,v)$ is $I(u,v)$, its integral image intensity $In(u,v)$ at the position of $(u,v)$ is defined as:

\begin{equation}
In(u,v)={\sum\limits_{i\le u, j\le v}I(i,j)}
\label{eq.integral_img_initialisation}
\end{equation}

Algorithm 1 details the implementation of the integral image initialisation, where $In$ is calculated serially based on its previous neighbouring results to minimise unnecessary computations.
\begin{algorithm}
	\SetKwInOut{Input}{Input}
	\SetKwInOut{Output}{Output}
	\Input{original image: $I$}
	\Output{integral image: $In$}
	$In(u_{min},v_{min})\gets I(u_{min},v_{min})$;\\
	\For{$u\gets u_{min}+1 $ to $u_{max}$}{
		$In(u,v_{min})\gets In(u-1,v_{min})+I(u,v_{min})$;\\
	}
	\For{$v\gets v_{min}+1 $ to $v_{max}$}{
		$In(u_{min},v)\gets In(u_{min},v-1)+I(u_{min},v)$;\\
	}
	\For{$u\gets u_{min}+1 $ to $u_{max}$}{
		\For{$v\gets v_{min}+1 $ to $v_{max}$}{
			$In(u,v)\gets In(u,v-1)+In(u-1,v)$\\$-In(u-1,v-1)+I(u,v)$;\\
		}
	}
	\caption{Integral image initialisation}
\end{algorithm}

After initialising an integral image, the sum $s(u,v)$ of pixel intensities within a square block whose side length is $2\rho+1$ and centre is $(u,v)$ can be computed with four references $r_1=In(u+\rho,v+\rho)$, $r_2=In(u-\rho-1,v-\rho-1)$, $r_3=In(u-\rho-1,v+\rho)$, and $r_4=In(u+\rho,v-\rho-1)$ as follows:
\begin{equation}
s(u,v)=r_1+r_2-r_3-r_4
\label{eq.integral_image}
\end{equation}

The mean $\mu(u,v)=s(u,v)/n$ of the intensities within the selected block is then stored in a static program storage for the computations of $\sigma$ and $c$. To simplify the computations of $\sigma_l$ and $\sigma_r$, we rearrange Eq. \ref{eq.sigma_l} and Eq. \ref{eq.sigma_r} as shown in Eq. \ref{eq.sigma_l_1} and Eq. \ref{eq.sigma_r_1}, respectively.

\begin{equation}
\sigma_{l}=\sqrt{\sum\limits_{i=u-\rho}^{i=u+\rho}\sum\limits_{j=v-\rho}^{j=v+\rho}{I_{l}}^2(i,j)/n-{\mu_{l}}^2}
\label{eq.sigma_l_1}
\end{equation}
\begin{equation}
\sigma_{r}=\sqrt{\sum\limits_{i=u-\rho}^{i=u+\rho}\sum\limits_{j=v-\rho}^{j=v+\rho}{I_{r}}^2(i-d,j)/n-{\mu_{r}}^2}
\label{eq.sigma_r_1}
\end{equation}
where $\sum{I_{l}}^2$ and $\sum {I_{r}}^2$ are dot products. Similarly, the computations of $\sum\ {I_{l}}^2$ and $\sum {I_{r}}^2$ can be accelerated by initialising two integral images $In_{{l^2}}$ and $In_{{r^2}}$ as references for indexing. Therefore, the standard deviations $\sigma_{l}$ and $\sigma_{r}$ can also be calculated and stored in static program storage for the efficient computation of $c$ as follows:
\begin{equation}
c(u,v,d)=\frac{1}{n\sigma_l \sigma_r}{\left[\sum\limits_{i=u-\rho}^{i=u+\rho}\sum\limits_{j=v-\rho}^{j=v+\rho} I_{l}(i,j) I_{r}(i-d,j)-n\mu_{l} \mu_{r}\right]}
\label{eq.ncc_1}
\end{equation}

From Eq. \ref{eq.ncc_1}, only $\sum I_{l}I_{r}$ needs to be calculated during the stereo matching. Hence, with the values of $\mu_l$, $\mu_r$, $\sigma_l$ and $\sigma_r$ able to be indexed directly, Eq. \ref{eq.ncc} is simplified as a dot product. The performance improvement achieved by factorising the NCC equation will be discussed section \ref{sec.st_evaluation}.

\begin{figure}[t!]
	\centering
	\subfigure[]
	{
		\includegraphics[width=0.46\textwidth]{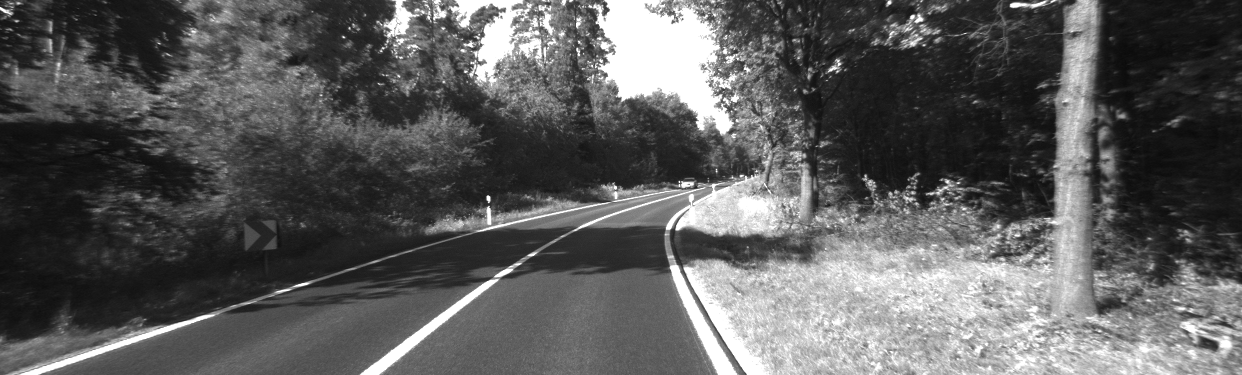}
		\label{fig.left}
	}	
	\subfigure[]
	{
		\includegraphics[width=0.46\textwidth]{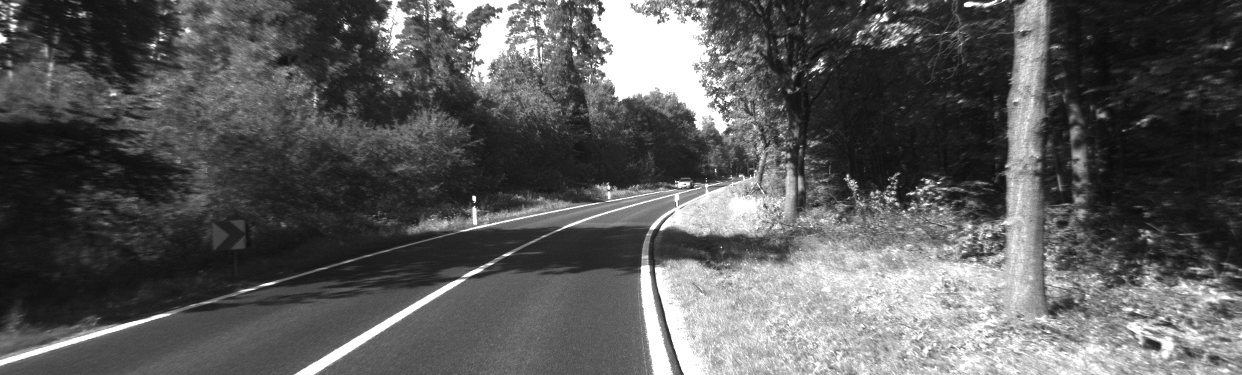}
		\label{fig.right}
	}
	\subfigure[]
	{
		\includegraphics[width=0.46\textwidth]{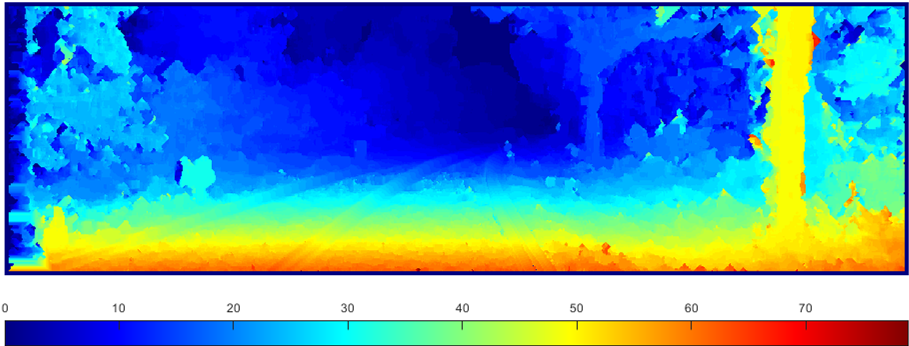}
		\label{fig.disp2}
	}
	\subfigure[]
	{
		\includegraphics[width=0.46\textwidth]{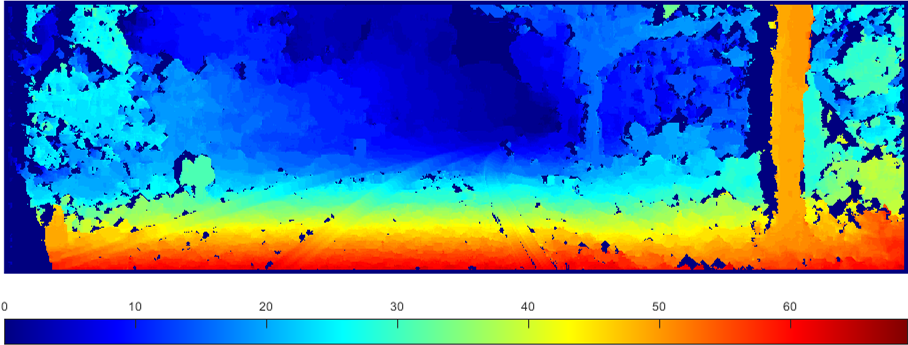}
		\label{fig.disp_lrc}
	}
	\caption{Input images and disparity maps. (a) left image. (b) right image. (c) disparity map. (d) disparity map processed with LRC. }
\end{figure}

\subsubsection{Search Range Propagation (SRP)}
\label{sec.srp}

In this paper, the disparities are estimated iteratively row by row from row $v_{max}$ to row $v_{min}$. In the first iteration, the stereo matching goes for a full search range $SR=\{sr|sr\in[d_{min},d_{max}]\}$. Then, the search range for stereo matching at the position of $(u,v)$ is propagated from three estimated neighbouring disparities $\ell(u-1,v+1)$, $\ell(u,v+1)$ and $\ell(u+1,v+1)$ using Eq. \ref{eq.srp} \cite{Fan2018}, where $\tau$ is the bound of the search range and set as $1$ in our proposed system. The left disparity map is illustrated in Fig. \ref{fig.disp2}. More details about the SRP-based disparity estimation are given in algorithm \ref{al.srp}. The performance of the SRP-based stereo will be discussed in section \ref{sec.st_evaluation}.


\begin{equation}
SR={\bigcup_{k=u-1}^{u+1}}\{sr|sr\in[\ell(k,v+1)-\tau,\ell(k,v+1)+\tau]\}
\label{eq.srp}
\end{equation}

\begin{algorithm}
	\SetKwInOut{Input}{Input}
	\SetKwInOut{Output}{Output}
	\Input{left image, right image;
		\\left mean map, right mean map;
		\\left standard deviation map, right standard deviation map;}
	\Output{disparity map}
	estimate the disparities for row $v_{max}$;\\
	\For{$v\gets v_{max}-1 $ to $v_{min}$}{
			\For{$u\gets u_{min} $ to $u_{max}$}{
		propagate the search range from row $v+1$ using Eq. \ref{eq.srp};\\
		estimate the disparity for $(u,v)$;\\
	}
	}
	\caption{SRP-based disparity map estimation}
	\label{al.srp}
\end{algorithm}

\subsubsection{Post-Processing}
\label{sec.st_post_processing}

For various disparity map estimation algorithms, the pixels that are only visible in one disparity map are a major source of the matching errors. Due to the uniqueness constraint of the correspondence, for an arbitrary pixel $(u,v)$ in the left disparity map $\ell^{lf}$, there exists at most one correspondence in the right disparity map $\ell^{rt}$, namely \cite{Fan2017}:
\begin{equation}
\ell^{lf}(u,v)=\ell^{rt}(u-\ell^{lf}(u,v),v)
\end{equation}

A left-right consistency (LRC) check is performed to remove half-occluded areas from the disparity map. Although the LRC check doubles the computational complexity by re-projecting the computed disparity values from one image to the other one, most of the incorrect half-occluded pixels can be eliminated and an outlier can be found \cite{Fan2017}. For $\ell^{rt}$ estimation, the memorisation of $\mu_{l}$, $\mu_{r}$, $\sigma_{l}$ and $\sigma_{r}$ is unnecessary because they have already been calculated when estimating $\ell^{lf}$. The LRC check is detailed in algorithm \ref{al.lrc}, where $tr_{LRC}$ is the threshold and set as 3. The corresponding results can be seen in Fig. \ref{fig.disp_lrc}.

\begin{algorithm}
		\SetKwInOut{Input}{Input}
	\SetKwInOut{Output}{Output}
	\Input{left disparity map: $\ell^{lf}$
		\\	right disparity map: $\ell^{rt}$}
	\Output{disparity map: $\ell$}
	
	\For{$v\gets v_{min} $ to $v_{max}$}{
		\For{$u\gets u_{min}$ to $u_{max}$}{
			\uIf{$abs(\ell^{lf}(u,v)-\ell^{rt}(u-\ell^{lf}(u,v),v))>tr_{LRC}$}{
				$\ell(u,v)\gets 0$;\\
			}
			\Else{
				$\ell(u,v)\gets \ell^{lf}(u,v)$;\\
			}
		}
	}
	\caption{LRC check}
	\label{al.lrc}
\end{algorithm}

\subsection{Dense $V_{py}$ Estimation}
\label{sec.vpy_estimation}

Since Labayrade et al. proposed the concept of "v-disparity" in 2002 \cite{Labayrade2002}, disparity information has been widely used to assist the detection of either obstacles or lanes. The v-disparity map is created by computing the histograms of each horizontal row of the disparity map. An example of the v-disparity map is shown in Fig. \ref{fig.v_disparity_map}, which has two axes: disparity $d$ and row number $v$. The value $m_{y}(d,v)$ represents the accumulation at the position of $(d, v)$ in the v-disparity map. In \cite{Hu2005}, Hu et al. proved that the projection of a flat road on the v-disparity map is a straight line: ${d}=f(v)=\alpha_0+\alpha_1v$. The parameters $\boldsymbol{\alpha}=[\alpha_0, \alpha_1]^\top$ can be obtained by using some linear pattern detectors such as the Hough Transform (HT) \cite{Ballard1981}. In our previous work \cite{Ozgunalp2017}, we used a parabola model ${d}=f(v)=\beta_0+\beta_1 v+\beta_2 v^2$ to depict the projection of a non-flat road surface on the v-disparity map. In this case, DP is more efficient than some quadratic pattern detectors to search for every possible solution and extract the projection path by minimising the energy function in Eq. \ref{eq.energy_function}.

\begin{equation}
\begin{split}
E=E_{data}+\lambda E_{smooth}
\end{split}
\label{eq.energy_function}
\end{equation}

Eq. \ref{eq.energy_function} is solved iteratively starting from $d=d_{max}$ and going to $d=0$. In the first iteration, $E_{smooth}=0$ and $E_{data}=-m_{y}(d_{max},v)$. Then, $E$ is computed based upon the previous iterations:

\begin{figure}[b!]
	\centering
	\subfigure[]
	{
		\includegraphics[width=0.14\textwidth]{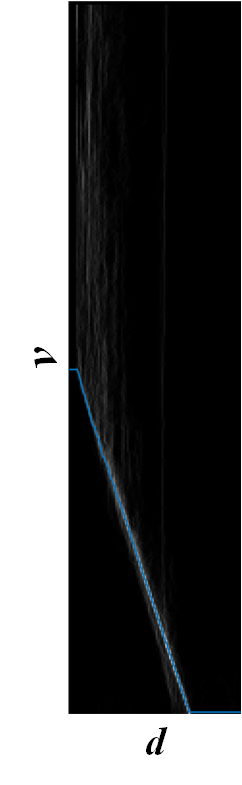}
		\label{fig.v_disparity_map}
	}
	\subfigure[]
	{
		\includegraphics[width=0.14\textwidth]{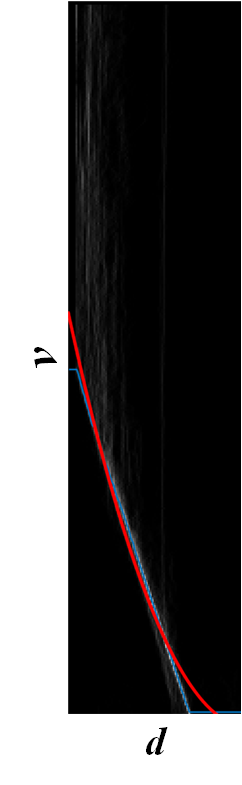}
		\label{fig.dynamic_programming_umar}
	}
	\subfigure[]
	{
		\includegraphics[width=0.14\textwidth]{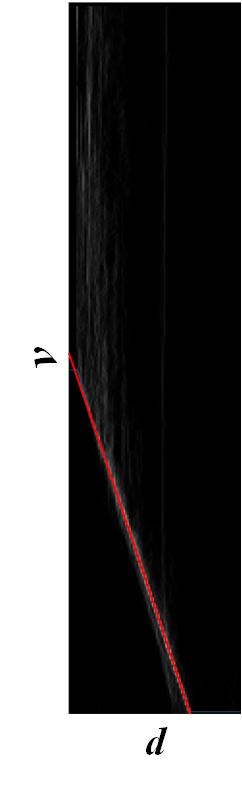}
		\label{fig.dynamic_programming_ranger}
	}
	\caption{V-disparity map and dynamic programming. (a) v-disparity map. (b) target solution in \cite{Ozgunalp2017}. (c) target solution in the proposed system. The blue paths are the optimal solution from DP. $f(v)=\beta_0+\beta_1v+\beta_2v^2$ is plotted in red.}
	\label{fig.v_disparity}
\end{figure}

\begin{equation}
\begin{split}
E(v)_d=-m_{y}(d,v) +\min_{\tau_y}[E(v-\tau_{y})_{d+1}-\lambda_{y}\tau_{y}],\ \text{s.t.} \  \tau_y\in[0,6]
\end{split}
\label{eq.vpy_estimation}
\end{equation}

In each iteration, the index position of the minimum is saved into a buffer for the solution back-tracing. The buffer has the same size as the v-disparity map. 
The solution $\boldsymbol{M_{y}}=[\boldsymbol{d}, \boldsymbol{v}]^\top\in\mathbb{R}^{k\times2}$ with the minimal energy is selected as the optima, which is plotted as the blue paths in Fig. \ref{fig.v_disparity}. The blue path has $k$ points. $\boldsymbol{v}=[v_0,v_1,\dots,v_{k-1}]^\top$ is a column vector recording their row numbers, and $\boldsymbol{d}=[d_0,d_1,\dots,d_{k-1}]^\top$ is a column vector storing their disparity values. Then, the parameters $\boldsymbol{\beta}=[\beta_0, \beta_1, \beta_2]^\top$ of the vertical road profile can be estimated by solving the least squares problem in Eq. \ref{eq.vpy_least_square_fitting}. We also plot $f(v)=\beta_0+\beta_1 v+\beta_2 v^2$ in red, as shown in Fig. \ref{fig.dynamic_programming_umar} and Fig. \ref{fig.dynamic_programming_ranger}.

\begin{equation}
\begin{split}
\boldsymbol{\beta}=\argminA_{\boldsymbol{\beta}} \sum_{j=0}^{k-1}(d_j-(\beta_0+\beta_1v_j+\beta_2{v_j}^2))^2  
\end{split}
\label{eq.vpy_least_square_fitting}
\end{equation}

From Fig.  \ref{fig.dynamic_programming_umar}, we observe that the outliers $\mathcal{O}$ severely affect the accuracy of the least squares fitting. To improve $V_{py}$ estimation, we employ RANSAC to update the inliers $\mathcal{I}$ and $\boldsymbol{\beta}$ iteratively. This procedure is described in algorithm \ref{al.ranger_dp_lsf_estimation}.

\begin{algorithm}
	\SetKwInOut{Input}{Input}
	\SetKwInOut{Output}{Output}
	\Input{optimal solution $\boldsymbol{M_{y}}=[\boldsymbol{d}, \boldsymbol{v}]^\top$}
	\Output{$\boldsymbol{\beta}$}
	\Do{$n_{\mathcal{I}}/(n_{\mathcal{I}}+n_{\mathcal{O}})<\epsilon_y$}{
	select a specified number of candidates $[d_j,v_j]^\top$ randomly\;
	fit a quadratic polynomial using Eq. \ref{eq.vpy_least_square_fitting} to get $\boldsymbol{\beta}$\;
	determine the number of inliers $\mathcal{I}$ and outliers $\mathcal{O}$: $n_{\mathcal{I}}$ and $n_{\mathcal{O}}$, respectively\;
	remove $\mathcal{O}$ from $\boldsymbol{M_{y}}$;
}

	fit $\mathcal{I}$ into a quadratic polynomial and get $\boldsymbol{\beta}$\;
	\caption{$\boldsymbol{\beta}$ estimation with the assist of RANSAC.}
	\label{al.ranger_dp_lsf_estimation}
\end{algorithm}

Given a candidate $[d_j,v_j]$, to decide whether it is an inlier or not, we need to compute the error $r_j=(d_j-f(v_j))^2$. If $r_j$ is smaller than our pre-set tolerance $tr_y$ ($tr_y=4$ in this paper), we mark this candidate as an inlier and update $\mathcal{I}$. Otherwise, it will be marked as an outlier and removed from $\boldsymbol{M_{y}}$. The iteration works until the fraction of inliers versus the number of candidates in the updated $\boldsymbol{M_{y}}$ exceeds our pre-set threshold $\epsilon_y$ ($\epsilon_y=0.99$ in this paper). Finally, the inliers are used to fit a quadratic polynomial to get $\boldsymbol{\beta}=[\beta_0,\beta_1,\beta_2]^\top$. Compared with the parabola obtained in \cite{Ozgunalp2017}, the parabola estimated with the assistance of RANSAC is more reliable and less affected by the outliers (please see Fig. \ref{fig.dynamic_programming_ranger}). $V_{py}$ can be computed as follows:

\begin{equation}
{V}_{py}{(v)}=v-\frac{\beta_0+\beta_1 v+\beta_2v^2}{\beta_1+2\beta_2v}
\label{eq.vpy_equation}
\end{equation}

\subsection{Dense $V_{px}$ Estimation}
\label{sec.dense_vpx_estimation}

\subsubsection{Sparse $V_{px}$ estimation}
\label{sec.sparse_vpx_estimation}

\begin{figure}[t!]
	\centering
	\subfigure[]
	{
		\includegraphics[width=0.46\textwidth]{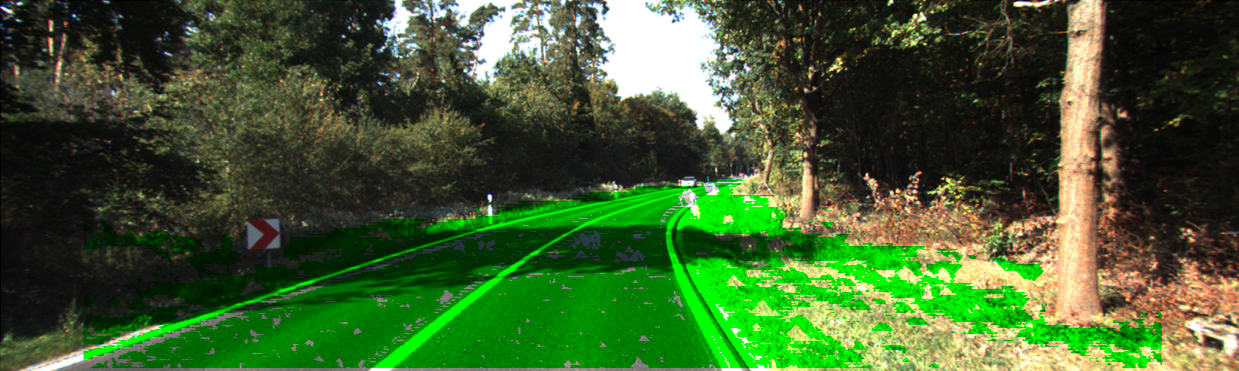}
		\label{fig.road_surface}
	}	
	\subfigure[]
	{
		\includegraphics[width=0.46\textwidth]{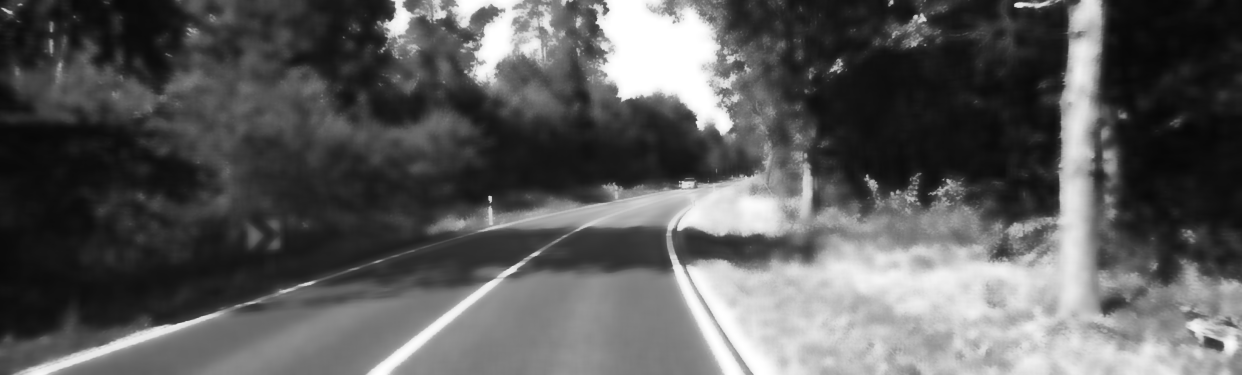}
		\label{fig.bilateral_output}
	}
	\subfigure[]
	{
		\includegraphics[width=0.46\textwidth]{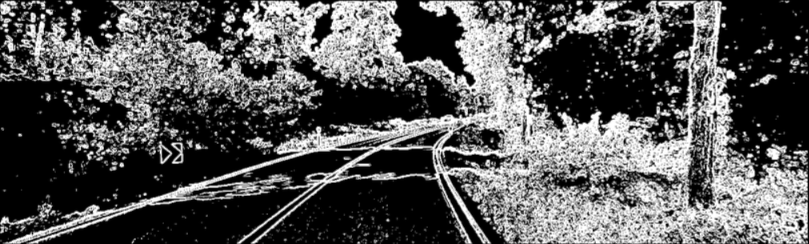}
		\label{fig.normal_edge}
	}
	\subfigure[]
	{
		\includegraphics[width=0.46\textwidth]{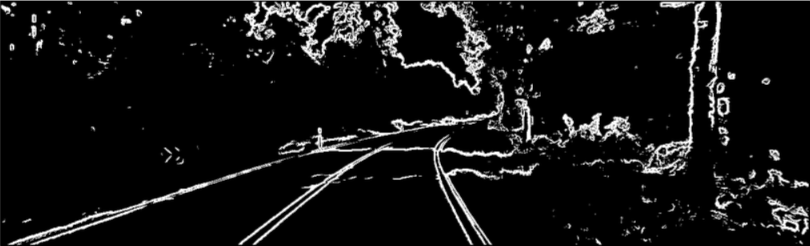}
		\label{fig.bilateral_edge}
	}
	\subfigure[]
	{
		\includegraphics[width=0.46\textwidth]{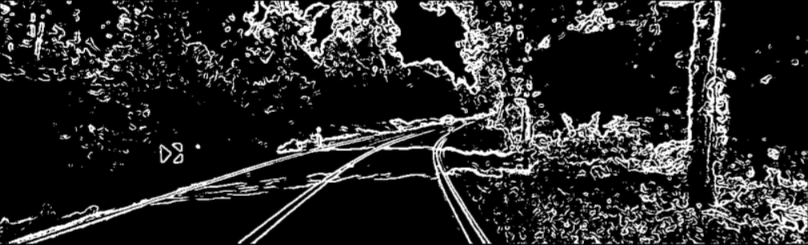}
		\label{fig.median_edge}
	}
	\subfigure[]
	{
		\includegraphics[width=0.46\textwidth]{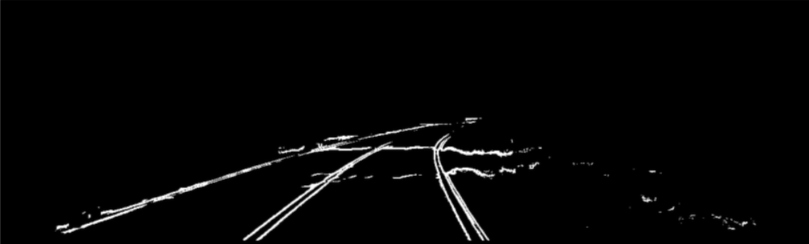}
		\label{fig.edge_map_noise_eliminated}
	}
	\caption{Sparse $V_{px}$ estimation. (a) road surface estimation. (b) bilateral filtering for Fig. \ref{fig.left}. (c) edge detection of Fig. \ref{fig.left}. (d) edge detection of (b). (e) edge detection of the median filtering output. (f) edges in the road surface area. The green area in (a) is the road surface. For the bilateral filter, $\sigma_s=300$ and $\sigma_r=0.3$. The window size of the bilateral filter and the median filter is $11\times11$. The thresholds of the Sobel edge detection in (c), (d), (e) and (f) are 100. For the following procedures, we only focus on the edge pixels in (f). }
\end{figure}

In order to reduce the redundant information, we set a threshold $\varpi=3$ to remove the non-road areas. A pixel at $(u,v)$ in the disparity map $\ell$ is identified to be on-road only if it satisfies $|\ell(u,v)-f(v)|\leq\varpi$ and $f^{-1}(0)\leq v \leq v_{max}$ ($f^{-1}$ is the inverse function of $f(v)=\beta_0+\beta_1 v+ \beta_2 v^2$). Otherwise, it is either on the obstacles or in the potholes. The estimated road surface is shown as a green area in Fig. \ref{fig.road_surface}.  For the following procedures, we only focus on the road surface area, which greatly reduces the redundant information used for edge detection and dense $V_{px}$ estimation. 

However, the noise introduced from the input image still makes the edge detectors like Sobel very sensitive to the blobs \cite{RafaelGonzalez2002}. Therefore, we use a bilateral filter to further reduce the noise before detecting edges. Compared with the median filter utilised in \cite{Ozgunalp2017}, the bilateral filter is more capable of preserving the edges when smoothing an image. The bilateral filtering \cite{He2013} is performed as follows:.
\begin{equation}
I^{bf}(u,v)=\frac{\sum\limits_{i=u-\rho}^{i=u+\rho}\sum\limits_{j=v-\rho}^{j=v+\rho}w_s(i, j)w_r(i,j)I(i, j)}{\sum\limits_{i=u-\rho}^{i=u+\rho}\sum\limits_{j=v-\rho}^{j=v+\rho}w_s(i,j)w_r(i,j)}
\label{eq.bilateral_filter}
\end{equation}
where 
\begin{equation}
\begin{split}
w_s(i,j)&= \exp \left\{ {-\frac{(i-u)^2+(j-v)^2}{\sigma_s^2}}\right\}\\
w_r(i,j)&= \exp \left\{ {-\frac{(I(i,j)-I(u,v))^2}{\sigma_r^2}}\right\}
\end{split}
\label{eq.bilateral_filter_ws_wr}
\end{equation}

$I(i,j)$ is the intensity of the input image at $(i,j)$ and $I^{bf}(u,v)$ is the intensity of the filtered image at $(u,v)$. The block size of the filter is $(2\rho+1)\times(2\rho+1)$, and its centre is $(u,v)$. The coefficient $w_s(i,j)$ is based on the spatial distance, and the coefficient $w_r(i,j)$ is based upon the colour similarity. $\sigma_s$ and $\sigma_r$ are the parameters for  $w_s$ and  $w_r$. In our practical experiments, $\sigma_s$ and $\sigma_r$ are set to 300 and 0.3, respectively. The output of bilateral filtering is shown in Fig. \ref{fig.bilateral_output}. The corresponding edge detection is illustrated as Fig. \ref{fig.bilateral_edge}. The edge detection of Fig. \ref{fig.left} is shown as Fig. \ref{fig.normal_edge}, while the edge detection for the output of the median filtering is depicted in Fig. \ref{fig.median_edge}. Obviously, although the median filter has removed a lot of redundant edges, the bilateral filter still achieves a better performance in terms of noise elimination and edge preservation. 

In the following procedures, we only focus on the pixels on the road surface. The edge map in the road surface area is shown in Fig. \ref{fig.edge_map_noise_eliminated}. Based on the gradient $\nabla(e)=[G_x,G_y]^\top$ approximated by the Sobel edge detector, we can estimate the sparse ${V}_{px}$ using Eq. \ref{eq.sparse_vpx_estimation} for every pixel $(u_e,v_e)$ on the edge $e$. ${V}^s_{px}(u_e,v_e)$ at the position of $(u_e,v_e)$ is recorded in a sparse $V_{px}$ map (note: we use the notation ${V}^s_{px}$ to represent the sparse ${V}_{px}$ in the following content).

\begin{equation}
{V}^s_{px}(u_e,v_e)=u_e+\frac{v_e-{V}_{py}(v_e)}{\nabla(e)}
\label{eq.sparse_vpx_estimation}
\end{equation}

Next, we will provide some details about the implementation. In the GPU architecture, a thread is more likely to fetch the memory from the closest addresses that its nearby threads accessed, which makes the use of cache impossible \cite{NVIDIA2017}. Therefore, we utilise the texture memory which is read-only and cached on-chip to optimise the caching for 2D spatial locality during the bilateral filtering. Firstly, a 2D texture object is created. Then, the texture object is bound directly to the global memory address of the left image. $I(i,j)$ used in the bilateral filtering is then fetched from the texture object to reduce the memory requests from the global memory. In addition, as the constant memory is read-only and beneficial for the data that will not change over the course of a kernel execution \cite{NVIDIA2017}, we create two lookup tables on it to store the values of $w_s$ and $w_r$. So far, the execution of bilateral filtering has been highly accelerated, so we move to the implementation of the Sobel edge detection. The address of $I^{bf}(u,v)$ is always accessed repeatedly when judging whether the pixel at $(u+\omega ,v+\varrho)$ belongs to an edge or not, where $\omega,\varrho\in\{-1,0,1\}$. Thus, we load a group of data $I^{bf}$ into the shared memory for each thread block.  All threads within the same thread block will access the shared data instead of fetching them repeatedly from the global memory. In order to avoid the race conditions among different threads which run logically in parallel instead of executing physically concurrently, the threads within the same thread block need to be synchronised after they finish the data loading. Compared with the performance of ${V}^s_{px}$ estimation on a Core-i7 4720HQ CPU processing with a single thread, our implementation on a GTX 970M GPU speeds up the execution by over $74$ times.

\subsubsection{Dense $V_{px}$ accumulation}
\label{sec.vpx_accumulation}
The sparse $V_{px}$ information is usually less robust than the dense $V_{px}$ information with regard to validating a correct lane position. 
Therefore, we propose to vote ${V}^s_{px}$ within a band whose longitudinal size is $2\chi+1$. The voting results are then saved in a 1D histogram. 
A 2D dense $V_{px}$ accumulator can be created by shifting an unfixed band from the bottom of the ${V}^s_{px}$ map to its top and gathering the 1D histogram from each band. 
We abbreviate dense $V_{px}$ as $V_{px}^d$ in the following content. 
The details of the $V_{px}^d$ accumulation are described in algorithm \ref{al.vpx_accumulation}.

\begin{algorithm}
	\SetKwInOut{Input}{Input}
	\SetKwInOut{Output}{Output}
	\Input{${V}^s_{px}$ map}
	\Output{$V_{px}^d$ accumulator}
	set all elements in $V_{px}^d$ accumulator as 0\;
	\For{$i\gets u_{min}$ \text{to} $u_{max}$}{$V_{px}^d({V}^s_{px}(i,v_{max}),v_{max})\gets-\varrho$}
	\For{$j\gets v_{max}-1$ \text{to} $f^{-1}(0)$}{
		\uIf{$j>v_{max}-\chi-1$}{
			\For{$i\gets u_{min}$ \text{to} $u_{max}$}{
				$V_{px}^d(i,j) \gets V_{px}^d(i,j+1)$\;
				$V_{px}^d({V}^s_{px}(i,j),j)\gets V_{px}^d({V}^s_{px}(i,j),j) -\varrho$\;
		}}
		\uElseIf{$f^{-1}(0)+\chi\leq j\leq v_{max}-\chi-1$}{
			\For{$i\gets u_{min}$ \text{to} $u_{max}$}{
				$V_{px}^d(i,j)\gets V_{px}^d(i,j+1)$\;
				$V_{px}^d(i,j)\gets V_{px}^d(i,j)+V_{px}^d(i,j+\chi+1)$\;
				$V_{px}^d({V}^s_{px}(i,j),j-\chi) \gets V_{px}^d({V}^s_{px}(i,j),j-\chi)-\varrho$\;
			}
		}
		\Else{
			\For{$i\gets u_{min}$ \text{to} $u_{max}$}{
				$V_{px}^d(i,j)\gets V_{px}^d(i,j+1)$\;
				$V_{px}^d(i,j)\gets V_{px}^d(i,j)+V_{px}^d(i,j+\chi+1)$\;
			}
		}
	}
	\caption{Dense $V_{px}$ accumulation.}
	\label{al.vpx_accumulation}
\end{algorithm}

\begin{figure}[t!]
	\centering
	\subfigure[]
	{
		\includegraphics[width=0.46\textwidth]{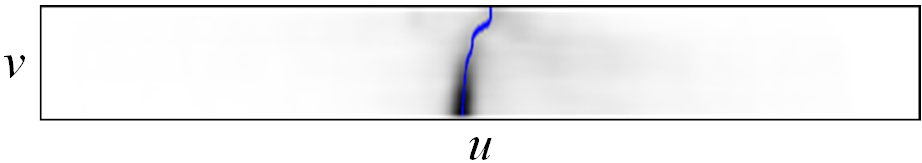}
		\label{fig.vpx_accumulator}
	}	
\\
	\subfigure[]
	{
		\includegraphics[width=0.46\textwidth]{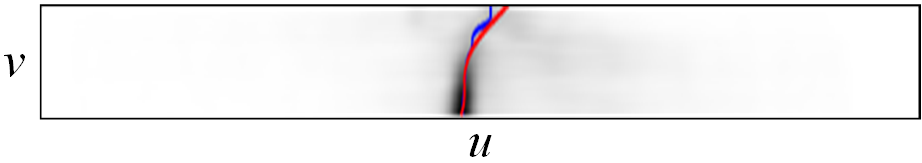}
		\label{fig.vpx_accumulator1}
	}
\\
	\subfigure[]
	{
		\includegraphics[width=0.46\textwidth]{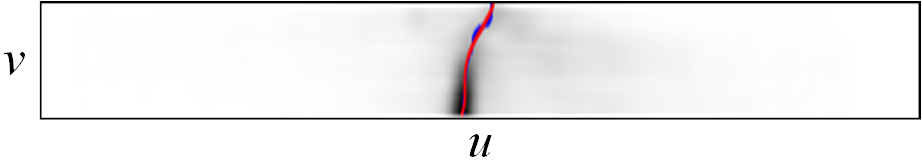}
		\label{fig.vpx_accumulator2}
	}
	\caption{Dense $V_{px}$ accumulator and dynamic programming. (a) dense $V_{px}$ accumulator. (b) target solution in \cite{Ozgunalp2017}. (c) target solution in the proposed system. The blue paths are the optimal solution obtained by DP. $g(v)=\gamma_0+\gamma_1v+\gamma_2v^2+\gamma_3v^3+\gamma_4v^4$ is plotted in red.}
	\label{fig.vpx}
\end{figure}

The initial step is to form a 1D $V_{px}^d$ histogram where each edge pixel $e$ on row $v_{max}$ votes ${V}^s_{px}$ to the histogram (the parameter $\varrho$ for each vote is proposed to be 1). Then, the votes of ${V}^s_{px}$ on row $v$ are accumulated with the 1D histogram on row $v+1$ to form the accumulation of $V^d_{px}$ on row $v$. This continues until the band is able to reach a $2\chi+1$ longitudinal size, where $\chi=25$ in this paper. 
Then, the current band is shifted slightly up to create another 1D $V_{px}^d$ histogram for statistics. In order to achieve the computational efficiency, a sliding window is used by simply subtracting the votes that appear on row $v-\chi$ above from the current band and adding the votes that appear on the bottom row of the previous band (Please note, we subtract the votes in order to ensure consistency with the term "energy minimisation" in DP, and therefore, more votes from the ${V}^s_{px}$ map correspond to a negatively higher value in the $V_{px}^d$ accumulator).
Furthermore, due to the higher lane curvature a road may have, a thinner band is more desirable for those top bands \cite{Ozgunalp2017}. Therefore, only the votes on row $v+\chi+1$ are added to them without the subtractions from the current band. The sliding window makes the 1D $V_{px}^d$ histogram update more efficiently by simply processing the bottom row and the top row. The 2D $V_{px}^d$ accumulator is illustrated in Fig. \ref{fig.vpx_accumulator}, where the notation $m_{x}(u,v)$ represents the votes of $V_{px}=u$ on row $v$. 


\subsubsection{$V_{px}$ estimation}
\label{sec.vpx_estimation}

Similarly, the $V^d_{px}$ accumulator is optimised by minimising the energy function in Eq. \ref{eq.energy_function} using the DP .  In the first iteration, $E_{smooth}=0$ and $E_{data}=m_{x}(u,v_{max})$. After that, $E$ is estimated based on the previous iterations:

\begin{equation}
\begin{split}
E(u)_v=m_{x}(u,v) +\min_{\tau_x}[E(V_{px}+\tau_x)_{v+1}+\lambda_{x}\tau_x] \ \text{s.t.} \  \tau_x\in[-5,5]
\end{split}
\label{eq.dense_vpx_estimation}
\end{equation}

The solution $\boldsymbol{M_{x}}=[\boldsymbol{u}, \boldsymbol{v}]\in\mathbb{R}^{t\times2}$ with the minimal energy is selected as the optima, and we plot it as a blue path in Fig. \ref{fig.vpx}. There are $t$ points on the path. $\boldsymbol{u}=[u_0,u_1,\dots,u_{t-1}]^\top$ is a column vector storing their column numbers, and $\boldsymbol{v}=[v_0,v_1,\dots,v_{t-1}]^\top$ is a column vector recording their row numbers. The parameters $\boldsymbol{\gamma}=[\gamma_0, \gamma_1, \gamma_2, \gamma_3, \gamma_4]^\top$ can be estimated by solving the least squares problem in Eq. \ref{eq.vpx_least_square_fitting}. Here, we use the same strategy as the estimation of $\boldsymbol{\beta}$. $\mathcal{I}$ and $\boldsymbol{M_{x}}$ are updated using RANSAC until their capacities do not change any more. Then, $\boldsymbol{\gamma}$ is estimated from $\mathcal{I}$.
Algorithm \ref{al.ranger_vpx_lsf_estimation} provides more details on the $\boldsymbol{\gamma}$ estimation.

\begin{equation}
\boldsymbol{\gamma}=\argminA_{\boldsymbol{\gamma}} \sum_{j=0}^{t-1}(u_j-(\gamma_0+\gamma_1v_j+\gamma_2{v_j}^2+\gamma_3{v_j}^3+\gamma_4{v_j}^4))^2
\label{eq.vpx_least_square_fitting}
\end{equation}

\begin{algorithm}[!h]
	\SetKwInOut{Input}{Input}
	\SetKwInOut{Output}{Output}
	\Input{The optimal solution $\boldsymbol{M_{x}}$}
	\Output{$\boldsymbol{\gamma}$}
	
		\Do{$n_{\mathcal{I}}/(n_{\mathcal{I}}+n_{\mathcal{O}})<\epsilon_x$}{
		select a specified number of candidates $[u_j,v_j]^\top$ randomly\;
		fit a quadruplicated polynomial using Eq. \ref{eq.vpx_least_square_fitting} to get $\boldsymbol{\gamma}$\;
		determine the number of inliers $\mathcal{I}$ and outliers $\mathcal{O}$: $n_{\mathcal{I}}$ and $n_{\mathcal{O}}$\;
		remove $\mathcal{O}$ from $\boldsymbol{M_{x}}$;
	}
	
	interpolate $\mathcal{I}$ into a quadruplicate polynomial and get $\boldsymbol{\gamma}$\;
	\caption{$\boldsymbol{\gamma}$ estimation with the assist of RANSAC.}
	\label{al.ranger_vpx_lsf_estimation}
\end{algorithm}

Given an observation data $[u_j,v_j]^\top$, to determine whether it is an inlier or outlier, the error $r_j=(u_j-g(v_j))^2$ will be computed, where the function $g(v)=\gamma_0+ \gamma_1v+ \gamma_2v^2+ \gamma_3v^3+ \gamma_4v^4$ is the solution of Eq. \ref{eq.vpx_least_square_fitting}.
If $r_j$ is smaller than our pre-set threshold $tr_x$ ($tr_x=16$ in this paper), the candidate $[u,v]^\top$ is marked as an inlier and saved in $\mathcal{I}$. Otherwise, it will be marked as an outlier $\mathcal{O}$ and removed from $\boldsymbol{M_{x}}$. The iteration runs until the fraction of $\mathcal{I}$ versus the capacity of $\boldsymbol{M_{x}}$ exceeds our pre-set threshold $\epsilon_x$ ($\epsilon_x=0.99$ in this paper). Finally, $\boldsymbol{\gamma}$ can be estimated from $\mathcal{I}$ by solving the energy minimisation problem in Eq. \ref{eq.vpx_least_square_fitting}. Compared with the target solution obtained in \cite{Ozgunalp2017}, as shown in Fig. \ref{fig.vpx_accumulator1}, the target solution in the proposed system is less affected by the outliers (please see Fig. \ref{fig.vpx_accumulator2}). In a practical implementation, we rearrange Eq. \ref{eq.vpx_least_square_fitting} as Eq. \ref{eq.vpx_fitting} to avoid the data overflow when fitting the quadruplicate polynomial.

\begin{equation}
\boldsymbol{\gamma}=(\kappa\boldsymbol{P}^\top \boldsymbol{P})^{-1}(\kappa\boldsymbol{P}^\top)\boldsymbol{u}
\label{eq.vpx_fitting}
\end{equation}
where 
\begin{equation}
\boldsymbol{P}=
\begin{bmatrix}
1 & {v_1} & \cdots & {v_1}^4\\
1 & {v_2} &  \cdots  & {v_2}^4\\
\vdots & \vdots & \ddots & \vdots\\
1 & {v_t} &  \cdots  & {v_t}^4\\
\end{bmatrix}
\label{eq.vandermonde_matrix}
\end{equation}

$\boldsymbol{P}$ is a Vandermonde matrix. $\kappa$ is used to avoid the data overflow problem caused by the higher order polynomials (e.g. when $v=375$, ${v}^8\approx4\times10^{20}$ which is far beyond the significand range of \textit{long double} type in C language). With $\boldsymbol{\gamma}$ estimated, the dense $V_{px}$ can thus be estimated for each row $v$ with the known parameters $\boldsymbol{\gamma}$ as follows:

\begin{equation}
V_{px}(v)=\gamma_0+\gamma_1v+\gamma_2{v}^2+\gamma_3{v}^3+\gamma_4{v}^4
\label{eq.vpx_estimation}
\end{equation}

\subsection{Lane Position Validation}
\label{sec.lane_validation_visualisation}

$\boldsymbol{{V}_p}$ provides the tangential direction and the curvature information of lanes, which can help us to validate the lane positions. In \cite{Ozgunalp2017}, we form a likelihood function $V(e)=\nabla(e)\cdot \cos(\theta_e-\theta_{\boldsymbol{V_p}})$ for each edge point $e$ to select the plus-minus peak pairs, where $V(e)$ is the vote from $e(u_e,v_e)$. $\theta_e$ is the angle between the $u$-axis and the orientation of $e$ which is approximated by a Sobel edge detector. $\theta_{\boldsymbol{V_p}}$ is the angle between the $u$-axis and the radial from an edge pixel $e$ to $\boldsymbol{V_p}(v_e)$. Although some impressive experimental results have been achieved by the local peak-pair selection algorithm in \cite{Ozgunalp2017}, some inaccurate detections are still inevitable. In this paper, we verify every possible solution efficiently using a global histogram $\mathcal{H}$ created by analysing the information of $G_x$ and $\boldsymbol{{V}_p}$. Then, the  favourable local minima are selected from $\mathcal{H}$ to determine the lane positions. 
Algorithm \ref{al.lane_position_validation} provides more details on the proposed approach.

\begin{algorithm}
	\SetKwInOut{Input}{Input}
	\SetKwInOut{Output}{Output}
	\Input{$\boldsymbol{V}_{p}$ and $G_x$\\}
	\Output{lane position vector $\boldsymbol{\Gamma }$}
	create two 2D maps $\mathcal{M}_0$, $\mathcal{M}_1$ with the same size of the input image $I$\;
	set all elements in $\mathcal{M}_0$, $\mathcal{M}_1$ as $0$\;
	create a 1D histogram $\mathcal{H}$ of size $(2\xi+1)u_{max}$\;
	$\forall\mathcal{M}_0(i,j)\gets \sum_{x=-\nu}^{x=+\nu} \sum_{y=-\varsigma}^{y=+\varsigma} G_x(i+x,j+y)w_g(i+x,j+y)$\;
	$\mathcal{M}_1\gets \mathcal{G}*\mathcal{M}_0$\;
	\For{$i\gets -\xi u_{max}$ \text{to} $\xi u_{max}$}{
		aggregate $\mathcal{M}_1$ from row $v_{max}$ to row $f^{-1}(0)$ to get the energy $E$\;	
		$\mathcal{H}(i+\xi u_{max})\gets E$\;	
	}
	\If{$\mathcal{H}(i)<\min\{ \mathcal{H}(i-1), \mathcal{H}(i+1)\}$}
	{
		put $i-\xi u_{max}$ into $\boldsymbol{\Gamma }$\;
	}	
	remove the elements which are smaller than the threshold $tr_{LPV}$  from $\boldsymbol{\Gamma}$\;
	remove the nearby candidates from $\boldsymbol{\Gamma}$\;
	multiple lanes visualisation\;
	\caption{Lane position validation.}
	\label{al.lane_position_validation}
\end{algorithm}

For a dark-light transition of a lane marking, $G_x$ is positive and higher than the values at the non-edge positions, while $-G_x$ is positive and higher on a light-dark transition of the lane markings \cite{Ozgunalp2017}. Therefore, our aim to validate a lane position turns out to be estimating the position of the middle between two adjacent peaks (dark-light transition and light-dark transition). In order to reduce the noise introduced from non-lane edges, we use a piecewise weighting $w_g$ to decrease their magnitudes:
\begin{equation}
w_{g}(u_e,v_e)=
\begin{cases}
\exp \left( {-\frac{|\theta_e-\theta_{\boldsymbol{V_p}}|}{{\sigma_g}^2}} \cdot \frac{36}{\pi}  \right),\quad |\theta_e-\theta_{\boldsymbol{V_p}}|\leq\frac{\pi}{6}\\
0,\qquad  \qquad \qquad \qquad \quad \quad \  \text{otherwise}\\
\end{cases} 
\label{eq.sigma_g}
\end{equation}
where we define the step as $\theta_s=\pi/36$. The portion $p=|\theta_e-\theta_{\boldsymbol{V_p}}|/\theta_s$ is used to provide a Gaussian weight so as to decrease the magnitude of noise pixels, where we set $\sigma_g=3.5$. Then, we accumulate the updated $G_x$ within a shifting box to further reduce the noise. The box size is $(2\nu+1)\times (2\varsigma+1)$, where $\nu=1$ and $\varsigma=3$ in our experiment. The accumulation output is saved in a 2D map $\mathcal{M}_0$, where the horizontal gradient from a dark-light peak to a light-dark peak is negative. To approximate the horizontal gradients, a convolution operation between the Sobel kernel $\mathcal{G}=[1,2,1]^\top[1,0,-1]$ and $\mathcal{M}_0$ is conducted and the output is stored in $\mathcal{M}_1$. Then, we aggregate the cost $\mathcal{M}_1$ for each possible position from row $v_{max}$ to row $f^{-1}(0)$:

\begin{equation}
E(v)_{u_{v}}=\mathcal{M}_1({u_{v}},v)+\lambda_g E(v+1)_{u_{v+1}}
\label{eq.lane_energy}
\end{equation}
where

\begin{equation}
u_v=\frac{{V}_{px}(v+1)+v u_{v+1}-V_{py}{(v+1)}u_{v+1}}{v+1-V_{py}}
\label{eq.lane_visualisation}
\end{equation}

In order to cover all possible solutions, we set $\xi=0.5$. This implies that $u$ starts from $-0.5 u_{max}$ and ends at $1.5u_{max}$. In the first iteration, we select a possible position $(u_{v_{max}},v_{max})$ and the total energy is simply represented as: $E=\mathcal{M}_1(u_{v_{max}},v_{max})$. Then, we use $\boldsymbol{V_p}$ to estimate the next position $(u_{v_{max}-1},v_{max}-1)$ on the selected track. The energy $E$ is updated using Eq. \ref{eq.lane_energy}, where $\lambda_g$ is proposed to be 1. The aggregation of $\mathcal{M}_1$ works until $v$ reaches $f^{-1}(0)$. For each lane starting from $(u_{v_{max}},v_{max})$, its total energy is saved in a 1D histogram $\mathcal{H}$. Then, we pick out the local minima which are smaller than our pre-set threshold $t$ from $\mathcal{H}$. At the same time, if two local minima are quite close to each other, we ignore the minima with a larger energy. Finally, the lanes can be visualised by iterating Eq. \ref{eq.lane_visualisation}. The lane detection results will be discussed in section \ref{sec.ld_evaluation}.

\section{Experimental Results}
\label{sec.ld_experimental_results}

\subsection{Stereo vision evaluation}
\label{sec.st_evaluation}

\begin{figure}[!b]
	\centering
	\includegraphics[width=1\textwidth]{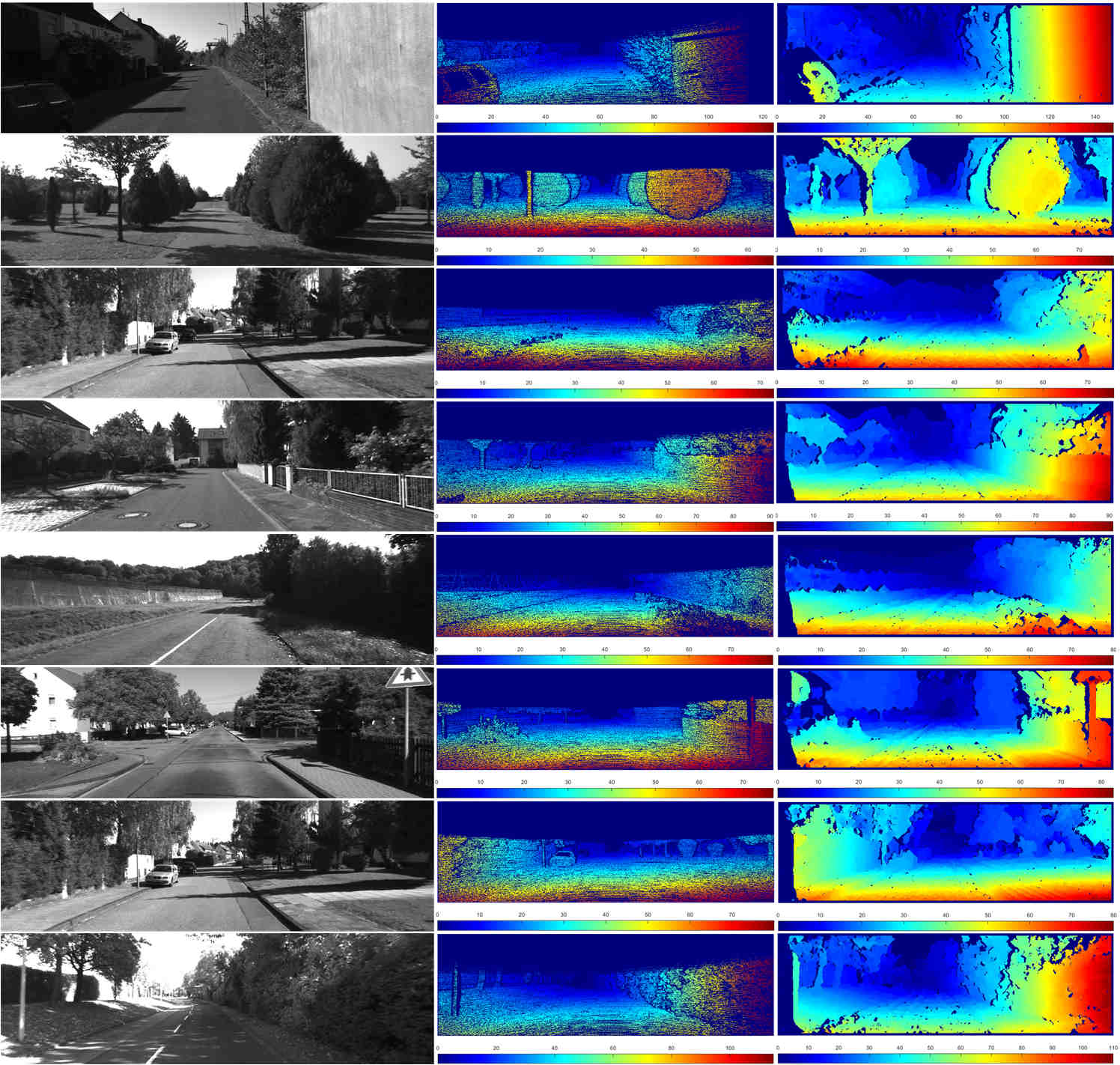}
	\caption{Experimental results of the KITTI stereo 2012 dataset.}
	\label{fig.stereo_vision_experimental_results}
\end{figure}

\begin{figure}[!t]
	\centering
	\includegraphics[width=0.5\textwidth]{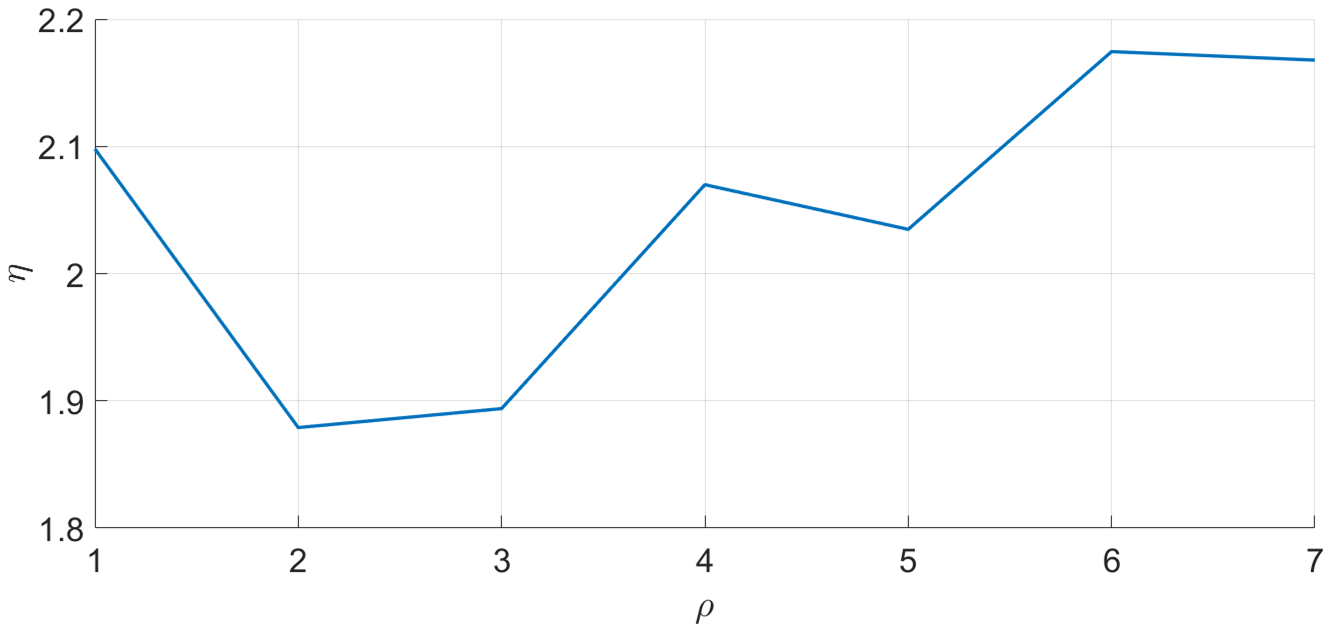}
	\caption{Evaluation of $\eta$ when using different $\rho$.}
	\label{fig.stereo_vision_eva}
\end{figure}
The proposed disparity estimation algorithm is evaluated using the KITTI stereo 2012 dataset \cite{Andreas2012}. Some experimental examples are shown in Fig. \ref{fig.stereo_vision_experimental_results}, where the first column illustrates the input left image, the second column shows the ground truth of the disparity map, and the third column depicts our experimental results. The overall percentage of the error pixels is approximately $6.82\%$ (error threshold: 2 pixels).

The proposed disparity estimation algorithm is implemented in C language on an i7-4720HQ CPU and a NVIDIA GTX 970M GPU for real-time purpose. Please kindly refer to \cite{Fan2017} for more details on the implementation. When $\rho=3$ and $\tau=1$, the algorithm execution achieves a speed of $37$ fps on GPU. After the memorisation, the values of $\mu$ and $\sigma$ are computed and stored in a static program storage for direct indexing in stereo matching, which greatly boosts the algorithm execution. The performance improvement achieved from memorisation is depicted in Fig. \ref{fig.stereo_vision_eva}, where $\eta$ represents the runtime of prevalent NCC-based stereo versus the runtime of NCC-based stereo with memorisation. In Fig. \ref{fig.stereo_vision_eva}, it can be clearly seen that the memorisation speeds up the algorithm execution by about two times.  

\subsection{Lane detection evaluation}
\label{sec.ld_evaluation}

\begin{figure}[!b]
	\centering
	\includegraphics[width=1\textwidth]{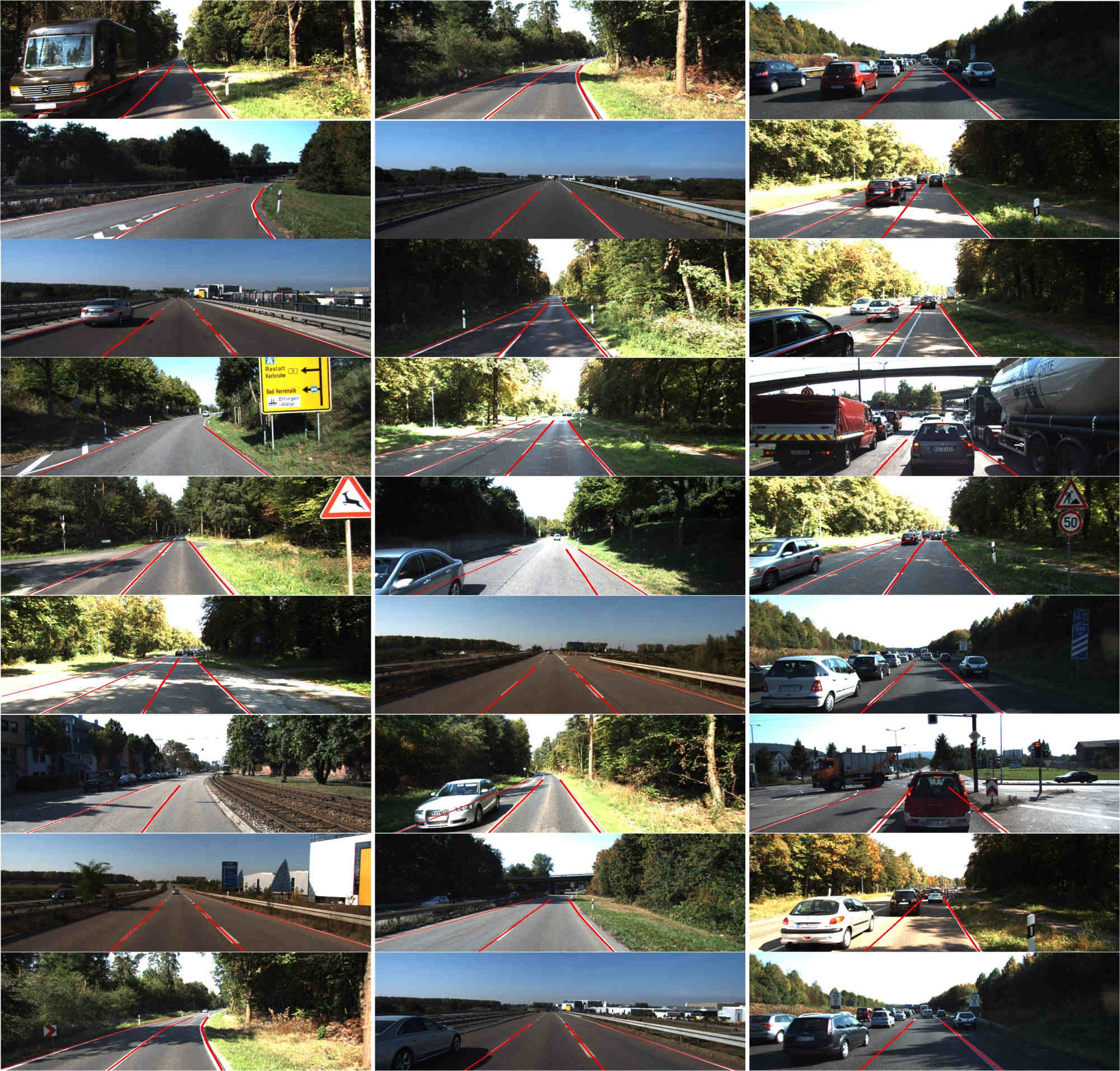}
	\caption{Lane detection experimental results. The red curves are the detected lanes. }
	\label{fig.experimental_results}
\end{figure}

Currently, it is impossible to access a satisfying ground truth dataset for the evaluation of lane detection algorithms because accepted test protocols do not usually exist \cite{Hillel2014}. Therefore, many publications for lane detection only focus on the quality of their experimental results. For this reason, we compare the performance of the proposed system with our previous work in \cite{Ozgunalp2017} in terms of both accuracy and speed. 
Some successful detection examples are shown in Fig. \ref{fig.experimental_results}.

\begin{table}[!h]
	\begin{center}
				\caption{Detection results of the proposed algorithm.}
		\label{table.proposed_algorithm_detection_results}
		\vspace{1em}
		\footnotesize
		\begin{tabular}{|c|c|c|c|}
			\hline
			Sequence  & Lanes  & Incorrect detection & Misdetection\\
			\hline
			1 & 860& 0 &  0 \\
			\hline
			2  & 594& 0 & 0 \\
			\hline
			3 & 376 &  0 & 0\\
			\hline
			4  & 156 &  0 & 0\\
			\hline
			5 & 678 &  0 & 0\\
			\hline
			6 & 1060 & 1 & 2\\
			\hline
			7  & 644  & 0 & 0\\
			\hline
			8 & 993  & 18 & 7\\
			\hline
			Total & 5361 & 19 & 9 \\			
			\hline
		\end{tabular}
	\end{center}
\end{table}

\begin{table}[!h]
	\begin{center}
				\caption{Detection results of \cite{Ozgunalp2017}.}
		\label{table.umar_detection_results}
		\vspace{1em}
		\footnotesize
		\begin{tabular}{|c|c|c|c|c|}
			\hline
			Sequence & Lanes  & Incorrect detection & Misdetection\\
			\hline
			1 & 860& 0 &  0 \\
			\hline
			2  & 594& 0 & 0 \\
			\hline
			3 & 376 &  0 & 0\\
			\hline
			4  & 156 &  0 & 9\\
			\hline
			5 & 678 &  0 & 17\\
			\hline
			6 & 1060 & 14 & 7\\
			\hline
			Total & 3724&  14 & 33 \\			
			\hline
		\end{tabular}
	\end{center}
\end{table}

\begin{figure}[t!]
	\centering
	\subfigure[]
	{
		\includegraphics[width=0.475\textwidth]{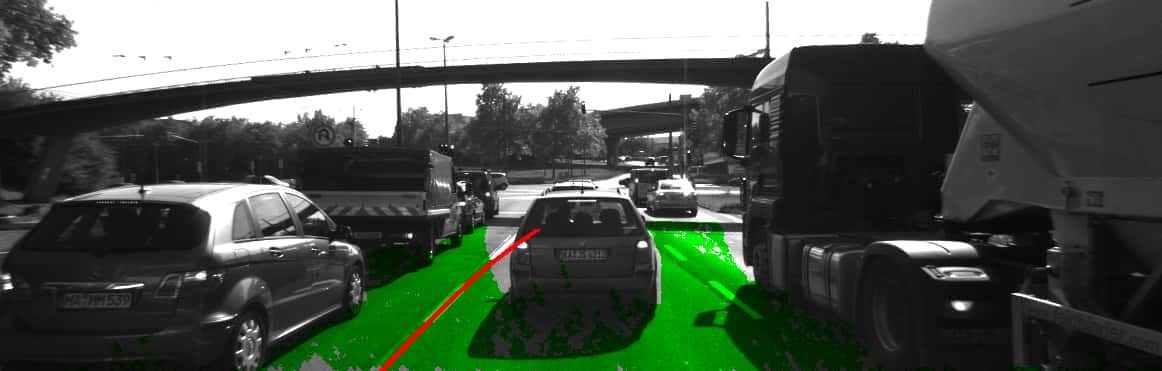}
		\label{fig.lane_umar1}
	}	
	\subfigure[]
	{
		\includegraphics[width=0.475\textwidth]{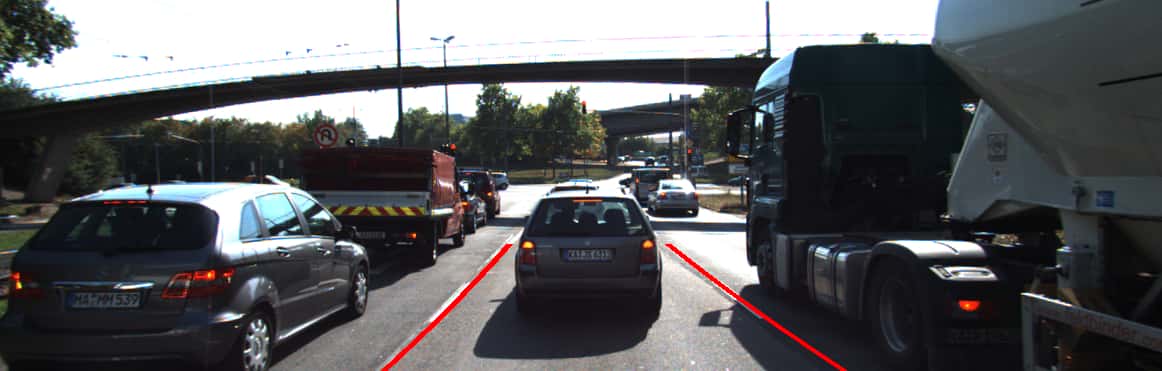}
		\label{fig.lane_ranger1}
	}
	
	\subfigure[]
	{
		\includegraphics[width=0.475\textwidth]{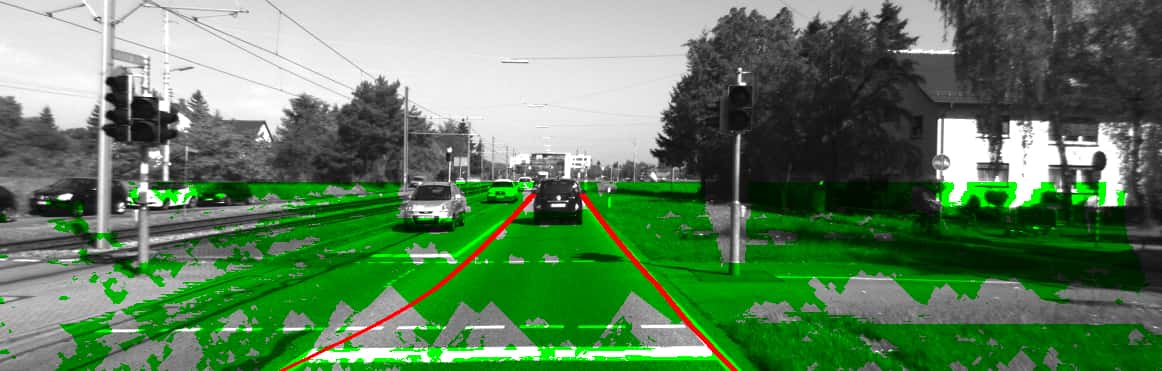}
		\label{fig.lane_umar4}
	}	
	\subfigure[]
	{
		\includegraphics[width=0.475\textwidth]{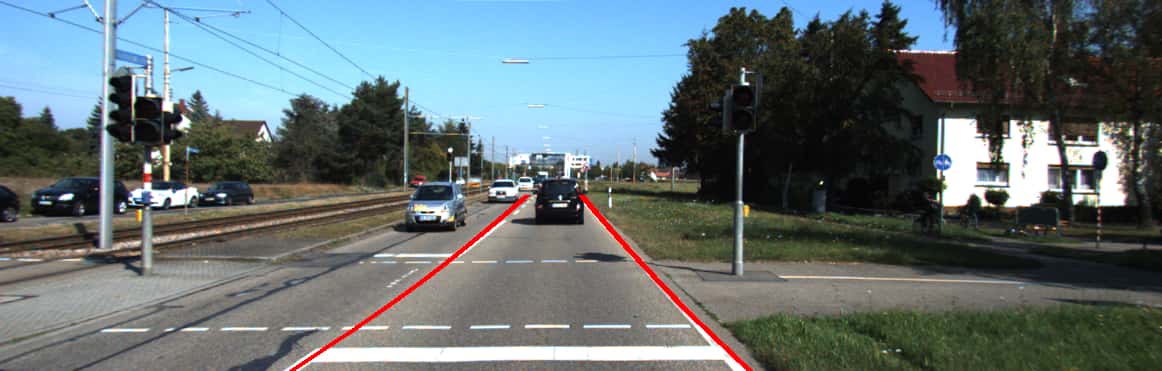}
		\label{fig.lane_ranger4}
	}	
	\subfigure[]
	{
		\includegraphics[width=0.475\textwidth]{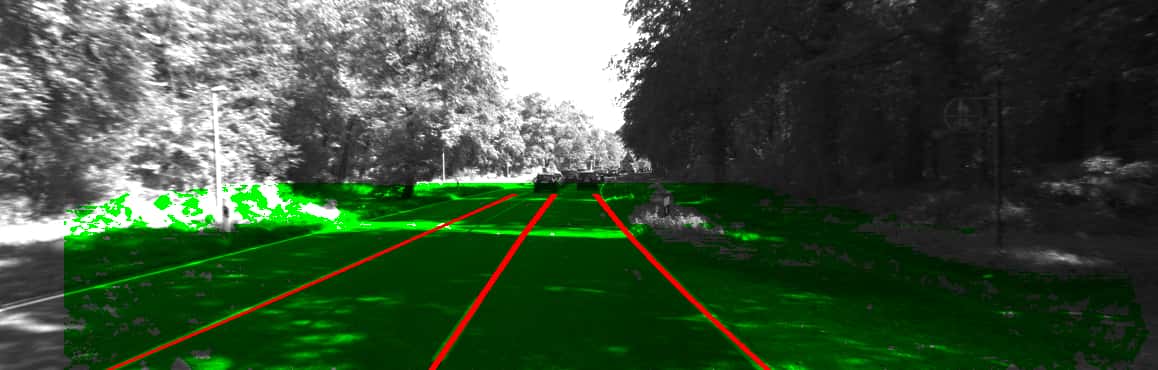}
		\label{fig.lane_umar2}
	}	
	\subfigure[]
	{
		\includegraphics[width=0.475\textwidth]{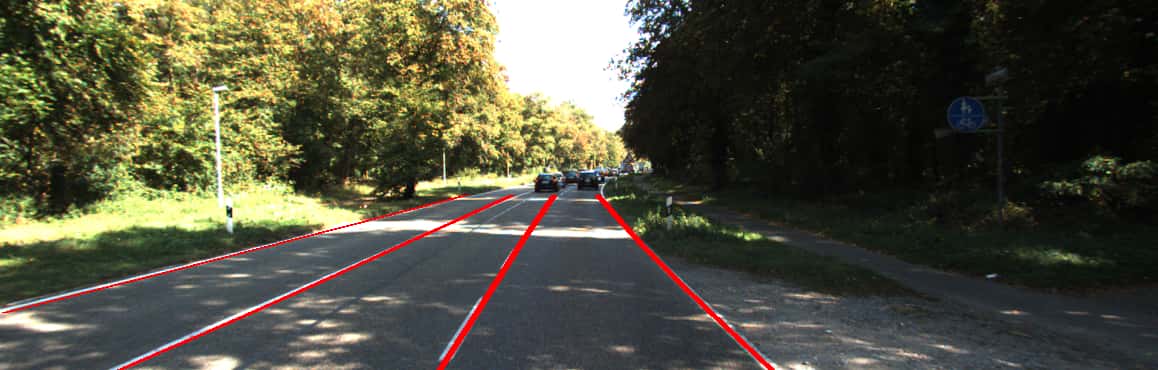}
		\label{fig.lane_ranger2}
	}
	\subfigure[]
	{
		\includegraphics[width=0.475\textwidth]{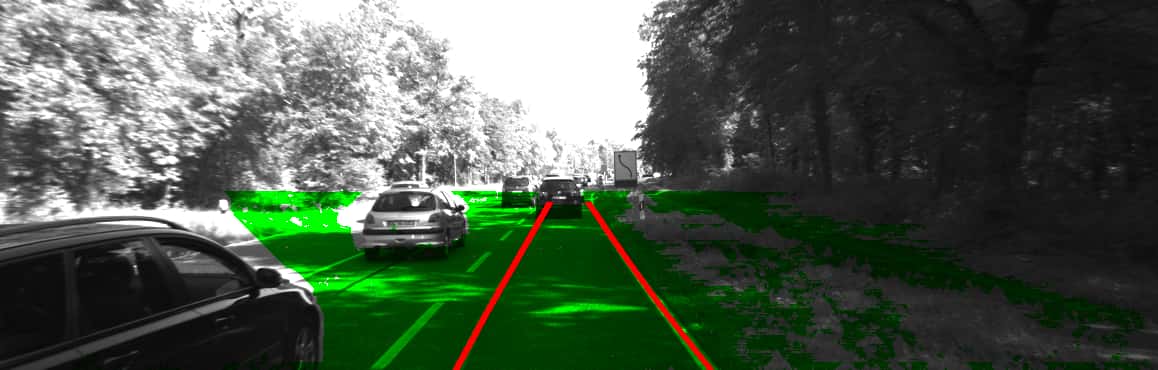}
		\label{fig.lane_umar3}
	}	
	\subfigure[]
	{
		\includegraphics[width=0.475\textwidth]{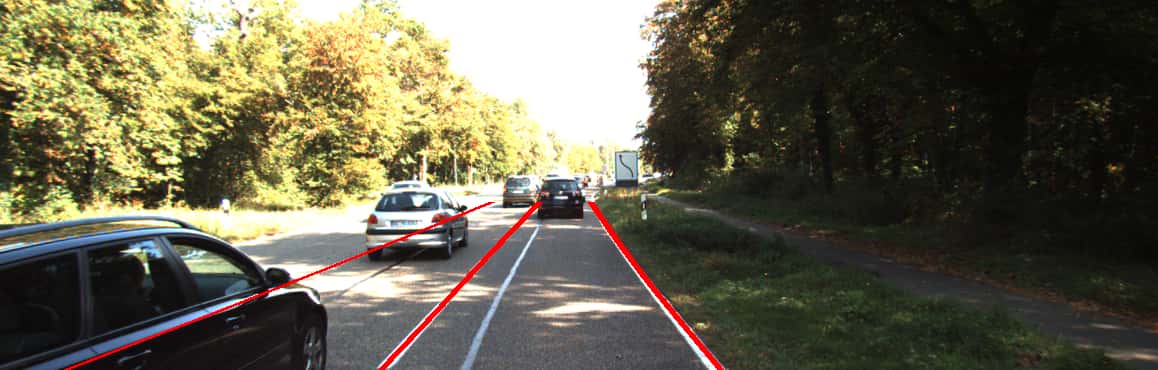}
		\label{fig.lane_ranger3}
	}
	
	\caption{Comparison between some failure examples in \cite{Ozgunalp2017} and the corresponding results in this paper. The green areas in the first column are the road surface. The red curves are the detected lanes. The first column illustrates the failed examples in \cite{Ozgunalp2017}, and the second column shows the corresponding results of the proposed system.}
	\label{fig.evaluation1}
	\vspace{-1.5em}
\end{figure}

Firstly, we will discuss the precision performance. The lane detection ratio of the proposed algorithm and our previous work \cite{Ozgunalp2017} are detailed in tables \ref{table.proposed_algorithm_detection_results} and \ref{table.umar_detection_results}, respectively. To evaluate the robustness of the proposed algorithm, we tested eight sequences (including two additional sequences): 2495 frames containing 5361 lanes from KITTI database \cite{Geiger2013} (1637 more lanes than the experiment in \cite{Ozgunalp2017}). The image resolution is $1242\times375$ in sequences 1 to 6, $1241\times376$ in sequence 7, and $1238\times374$ in sequence 8. From table \ref{table.proposed_algorithm_detection_results}, it can be seen that the proposed algorithm presents a better detection ratio where $99.9\%$ lanes are successfully detected in sequences 1 to 7 (including all the sequences in table \ref{table.umar_detection_results}), while the detection ratio of \cite{Ozgunalp2017} is $98.7\%$. The comparison between some failed examples in \cite{Ozgunalp2017} and their corresponding results of the proposed system is illustrated in Fig. \ref{fig.evaluation1}.

In Fig. \ref{fig.lane_umar1},  we can see that the obstacle areas occupy a larger portion than the road surface area, which severely affects the accuracy of $V_{py}$. When we consider both inliers and outliers into the LSF, $\boldsymbol{{V}_p}$ differs too much from the ground truth. This further influences the peak-pair selection and leads to an imprecise detection and a misdetection.  When reflecting on Fig. \ref{fig.lane_ranger1}, it can be seen that the fitting with only inliers increases the precision of the $V_{py}$ estimation significantly. Moving to the second row, an over-curved lane can be seen in Fig. \ref{fig.lane_umar4}. When we fit $\boldsymbol{\beta}$ and $\boldsymbol{\gamma}$ with the assist of RANSAC, the improvement can be observed in Fig. \ref{fig.lane_ranger4}, where a more reasonable lane is detected. In Fig. \ref{fig.lane_umar2}, the lane near the left road boundary is misdetected because the low contrast between lane and road surface reduces its magnitude in the peak-pair selection stage. In Fig. \ref{fig.lane_umar3}, an incorrect detection occurs because the magnitude of a plus-minus peak of road markings is higher than the magnitude of lanes. In section \ref{sec.lane_validation_visualisation}, we proposed a more effective piecewise weighting $w_g$ to update $G_x$ for the edge pixels. Then, $G_x$ of the non-lane edges reduces significantly, which therefore greatly helps us avoid the incorrect detection of some lane markings. Also, we sum $G_x w_g$ within a shifting box for each position, which increases the magnitude of the lanes which are lowly contrastive to the road surface. The misdetections in Fig. \ref{fig.lane_umar2} and \ref{fig.lane_umar3} are thus detectable, and the failed detection in \ref{fig.lane_umar3} is also corrected. The corresponding results of the proposed system are shown in Fig. \ref{fig.lane_ranger2} and Fig. \ref{fig.lane_ranger3}.

\begin{figure}[b!]
	\centering
	\subfigure[]
	{
		\includegraphics[width=0.475\textwidth]{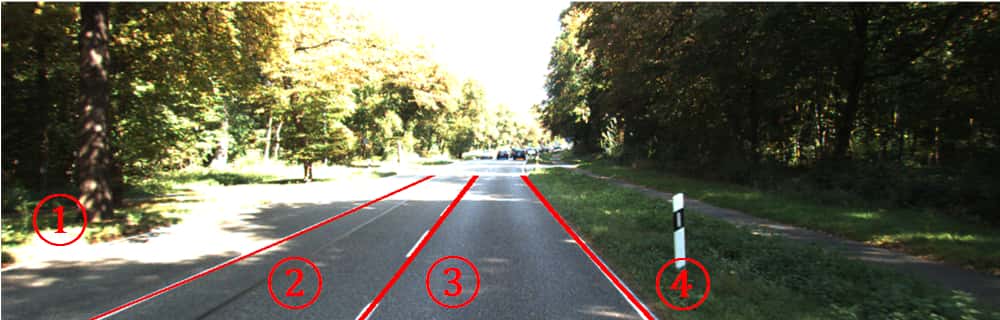}
		\label{fig.lane_detection_failure1}
	}	
	\subfigure[]
	{
		\includegraphics[width=0.475\textwidth]{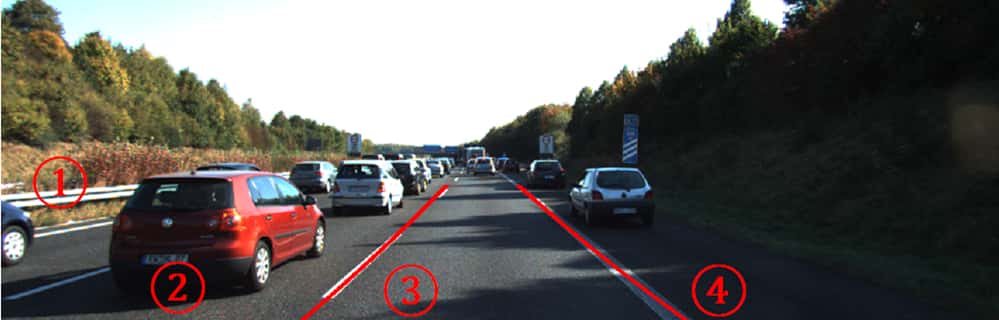}
		\label{fig.lane_detection_failure5}
	}
	\subfigure[]
	{
		\includegraphics[width=0.475\textwidth]{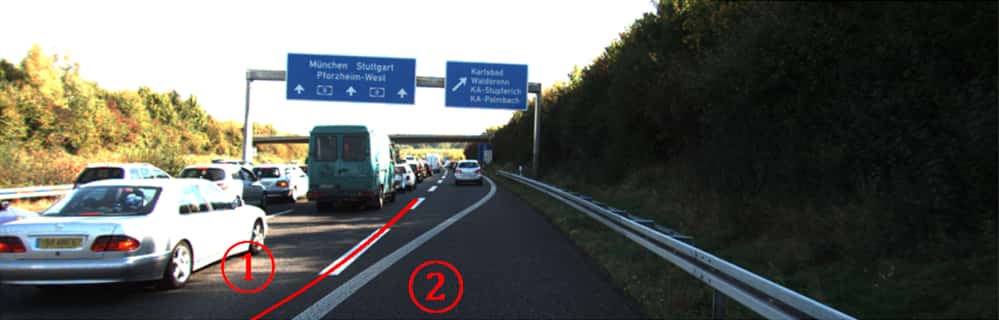}
		\label{fig.lane_detection_failure4}
	}
	\subfigure[]
	{
		\includegraphics[width=0.475\textwidth]{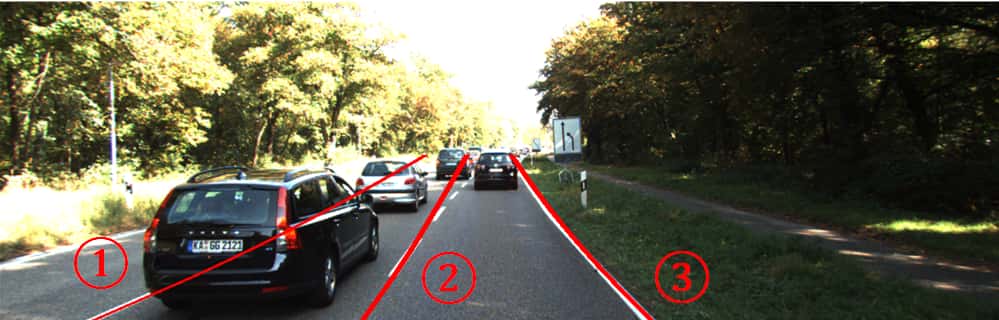}
		\label{fig.lane_detection_failure2}
	}	
	\subfigure[]
	{
		\includegraphics[width=0.475\textwidth]{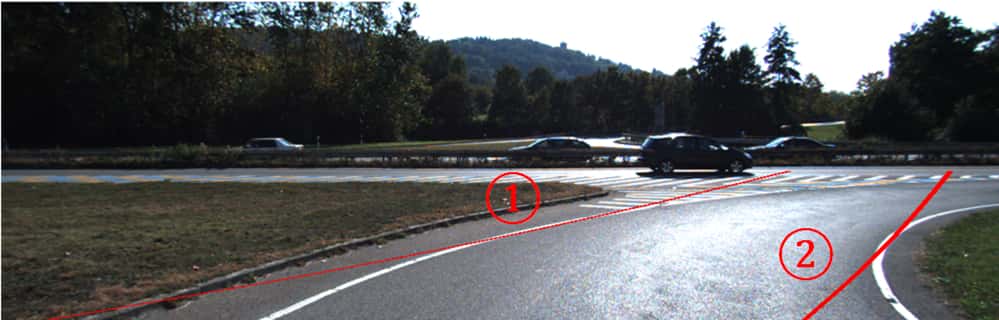}
		\label{fig.lane_detection_failure3}
	}
	\subfigure[]
	{
		\includegraphics[width=0.475\textwidth]{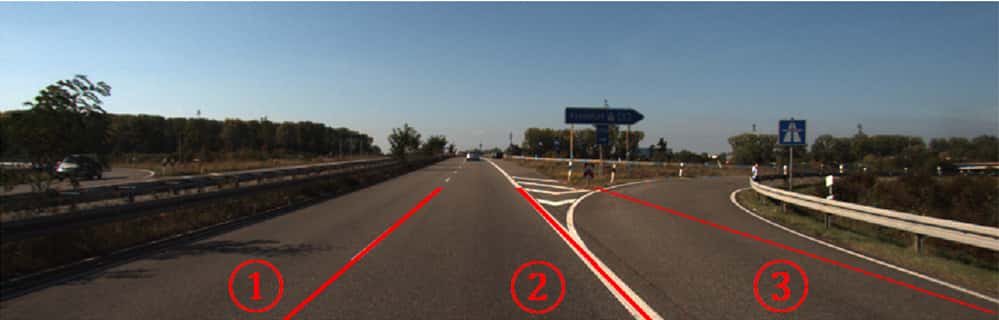}
		\label{fig.lane_detection_failure6}
	}	
	\caption{Examples of the failed detection in the proposed algorithm.}
	\label{fig.lane_failure}
\end{figure}

In our experiment, the failed cases consist of misdetections and incorrect detections. The misdetections are mainly caused by: the over-exposure of the image, partially-occluded by the obstacles, forks on the road. The corresponding examples are illustrated in Fig. \ref{fig.lane_detection_failure1}, \ref{fig.lane_detection_failure5} and \ref{fig.lane_detection_failure4}, respectively. In Fig. \ref{fig.lane_detection_failure1}, due to the image over-exposure, the edges pixels on lane 1 are rare, which leads to its misdetection. In Fig. \ref{fig.lane_detection_failure5}, we can see that the vehicles partially occlude lane 1 and lane 2. The occlusion decreases the magnitudes of $G_x w_g$ when we try to validate the lane positions, which makes lane 1 and lane 2 undetectable. In \ref{fig.lane_detection_failure4}, lane 2 forks from lane 1, and thus, has a different curvature information from lane 1, which therefore causes the misdetection.

For the factors leading to the incorrect detections, we group them into three main categories: ambiguous projection of road surface on the v-disparity map; $\boldsymbol{{V}_p}$ that does not exist in the image; different roadways, which are presented in Fig. \ref{fig.lane_detection_failure2},  \ref{fig.lane_detection_failure3} and \ref{fig.lane_detection_failure6}, respectively. In Fig. \ref{fig.lane_detection_failure2}, the obstacles (i.e. vehicles, trees and signageS) take a big portion in the image. Therefore, when $d$ is around 0, $m_y$ is mainly accumulated by the pixels on the obstacles and sky, which makes $V_{py}$ inaccurate and lanes 1 to 3 are slightly above the ground truth when the lane moves to the boundary between the road surface and sky. In Fig. \ref{fig.lane_detection_failure3}, $\boldsymbol{V_p}$ does not exist, which affects the detection results. In Fig. \ref{fig.lane_detection_failure6}, there are two different roadways: roadway between lane 1 and lane 2; roadway between lane 2 and lane 3. The second roadway turns right and therefore has a different $\boldsymbol{{V}_p}$ from the first roadway, which leads to an imprecise detection of lane 3.

Next, we will discuss about the speed performance. The algorithm is implemented on a heterogeneous system consisting of an Intel Core i7-4720HQ CPU and an NVIDIA GTX 970M GPU. The GPU has 10 streaming multiprocessors (SMs) with 128 CUDA cores on each of them. In the sparse $V_{px}$ estimation stage, the optimisations of the GPU memory allocation make this stage yield a performance speed-up of over 74 times than the implementation on a CPU with single thread processing. 
The total runtime of the proposed system is around 7 ms, which is approximately 38 times faster than our previous work where 263 ms was achieved (excluding the runtime of the disparity estimation). The authors believe that the failed cases can be prevented in the future with the additional of a lane tracking algorithm. A sample video is available at https://www.youtube.com/watch?v=fgriUdy1kv0.

\section{Conclusion}
\label{sec.ld_conclusion}
A multiple lane detection system was presented in this paper. The novelties in this paper include: improved dense $\boldsymbol{{V}_p}$ estimation using RANSAC, image pre-processing with the bilateral filtering, lane position validation with $G_x$ and $\boldsymbol{{V}_p}$, and a real-time implementation on a CPU-GPU-based heterogeneous system. 
To evaluate the performance, 5361 lanes from eight datasets were tested. 
The experimental results illustrated that the proposed algorithm works more accurately and robustly than our previous work with a $99.5\%$ successful detection ratio achieved. By highly exploiting the GPU architecture and allocating different parts on different platforms for execution, a high processing speed of 143 fps is achieved, which is approximately 38 times faster than our previous work in \cite{Ozgunalp2017}.

\vspace{3em}
\bibliographystyle{IEEEbib}

\begin{thebibliography}{10}
	
	\bibitem{Rosenzweig2015}
	Juan Rosenzweig and Michael Bartl,
	\newblock ``A review and analysis of literature on autonomous driving,''
	\newblock {\em E-Journal Making-of Innovation}, 2015.
	
	\bibitem{Narote2018}
	Sandipann~P Narote, Pradnya~N Bhujbal, Abbhilasha~S Narote, and Dhiraj~M Dhane,
	\newblock ``A review of recent advances in lane detection and departure warning
	system,''
	\newblock {\em Pattern Recognition}, vol. 73, pp. 216--234, 2018.
	
	\bibitem{Bertozzi1998}
	Massimo Bertozzi and Alberto Broggi,
	\newblock ``Gold: A parallel real-time stereo vision system for generic
	obstacle and lane detection,''
	\newblock {\em IEEE transactions on image processing}, vol. 7, no. 1, pp.
	62--81, 1998.
	
	\bibitem{Wang2004}
	Yue Wang, Eam~Khwang Teoh, and Dinggang Shen,
	\newblock ``Lane detection and tracking using b-snake,''
	\newblock {\em Image and Vision computing}, vol. 22, no. 4, pp. 269--280, 2004.
	
	\bibitem{Fan2016}
	Rui Fan, Victor Prokhorov, and Naim Dahnoun,
	\newblock ``Faster-than-real-time linear lane detection implementation using
	soc dsp tms320c6678,''
	\newblock in {\em Imaging Systems and Techniques (IST), 2016 IEEE International
		Conference on}. IEEE, 2016, pp. 306--311.
	
	\bibitem{Ozgunalp2014}
	Umar Ozgunalp and Naim Dahnoun,
	\newblock ``Robust lane detection \& tracking based on novel feature extraction
	and lane categorization,''
	\newblock in {\em Acoustics, Speech and Signal Processing (ICASSP), 2014 IEEE
		International Conference on}. IEEE, 2014, pp. 8129--8133.
	
	\bibitem{Kluge1995}
	Karl Kluge and Sridhar Lakshmanan,
	\newblock ``A deformable-template approach to lane detection,''
	\newblock in {\em Intelligent Vehicles' 95 Symposium., Proceedings of the}.
	IEEE, 1995, pp. 54--59.
	
	\bibitem{Wang2008}
	Yan Wang, Li~Bai, and Michael Fairhurst,
	\newblock ``Robust road modeling and tracking using condensation,''
	\newblock {\em IEEE Transactions on Intelligent Transportation Systems}, vol.
	9, no. 4, pp. 570--579, 2008.
	
	\bibitem{Zhou2006}
	Yong Zhou, Rong Xu, Xiaofeng Hu, and Qingtai Ye,
	\newblock ``A robust lane detection and tracking method based on computer
	vision,''
	\newblock {\em Measurement science and technology}, vol. 17, no. 4, pp. 736,
	2006.
	
	\bibitem{Kreucher1999}
	Chris Kreucher and Sridhar Lakshmanan,
	\newblock ``Lana: a lane extraction algorithm that uses frequency domain
	features,''
	\newblock {\em IEEE Transactions on Robotics and automation}, vol. 15, no. 2,
	pp. 343--350, 1999.
	
	\bibitem{Jung2005}
	Cl{\'a}udio~Rosito Jung and Christian~Roberto Kelber,
	\newblock ``An improved linear-parabolic model for lane following and curve
	detection,''
	\newblock in {\em Computer Graphics and Image Processing, 2005. SIBGRAPI 2005.
		18th Brazilian Symposium on}. IEEE, 2005, pp. 131--138.
	
	\bibitem{Wang2000}
	Yue Wang, Dinggang Shen, and Eam~Khwang Teoh,
	\newblock ``Lane detection using spline model,''
	\newblock {\em Pattern Recognition Letters}, vol. 21, no. 8, pp. 677--689,
	2000.
	
	\bibitem{Nieto2007}
	Marcos Nieto, Luis Salgado, Fernando Jaureguizar, and Julian Cabrera,
	\newblock ``Stabilization of inverse perspective mapping images based on robust
	vanishing point estimation,''
	\newblock in {\em Intelligent Vehicles Symposium, 2007 IEEE}. IEEE, 2007, pp.
	315--320.
	
	\bibitem{Schreiber2005}
	David Schreiber, Bram Alefs, and Markus Clabian,
	\newblock ``Single camera lane detection and tracking,''
	\newblock in {\em Intelligent Transportation Systems, 2005. Proceedings. 2005
		IEEE}. IEEE, 2005, pp. 302--307.
	
	\bibitem{Hanwell2012}
	David Hanwell and Majid Mirmehdi,
	\newblock ``Detection of lane departure on high-speed roads.,''
	\newblock in {\em ICPRAM (2)}, 2012, pp. 529--536.
	
	\bibitem{Fardi2004}
	Basel Fardi and Gerd Wanielik,
	\newblock ``Hough transformation based approach for road border detection in
	infrared images,''
	\newblock in {\em Intelligent Vehicles Symposium, 2004 IEEE}. IEEE, 2004, pp.
	549--554.
	
	\bibitem{Wang2012}
	Yifei Wang, Naim Dahnoun, and Alin Achim,
	\newblock ``A novel system for robust lane detection and tracking,''
	\newblock {\em Signal Processing}, vol. 92, no. 2, pp. 319--334, 2012.
	
	\bibitem{Ozgunalp2017}
	Umar Ozgunalp, Rui Fan, Xiao Ai, and Naim Dahnoun,
	\newblock ``Multiple lane detection algorithm based on novel dense vanishing
	point estimation,''
	\newblock {\em IEEE Transactions on Intelligent Transportation Systems}, vol.
	18, no. 3, pp. 621--632, 2017.
	
	\bibitem{Labayrade2002}
	Raphael Labayrade, Didier Aubert, and J-P Tarel,
	\newblock ``Real time obstacle detection in stereovision on non flat road
	geometry through" v-disparity" representation,''
	\newblock in {\em Intelligent Vehicle Symposium, 2002. IEEE}. IEEE, 2002,
	vol.~2, pp. 646--651.
	
	\bibitem{Zhang2013}
	Zhen Zhang, Xiao Ai, and Naim Dahnoun,
	\newblock ``Efficient disparity calculation based on stereo vision with ground
	obstacle assumption,''
	\newblock in {\em Signal Processing Conference (EUSIPCO), 2013 Proceedings of
		the 21st European}. IEEE, 2013, pp. 1--5.
	
	\bibitem{Fan2017}
	Rui Fan and Naim Dahnoun,
	\newblock ``Real-time implementation of stereo vision based on optimised
	normalised cross-correlation and propagated search range on a gpu,''
	\newblock in {\em Imaging Systems and Techniques (IST), 2017 IEEE International
		Conference on}. IEEE, 2017, pp. 241--246.
	
	\bibitem{Dahnoun}
	Naim Dahnoun,
	\newblock ``Stereo vision implementation,''
	\newblock {\em Multicore DSP: From Algorithms to Real-time Implementation on
		the TMS320C66x SoC}, pp. 604--616.
	
	\bibitem{Lewis1995}
	John~P Lewis,
	\newblock ``Fast template matching,''
	\newblock in {\em Vision interface}, 1995, vol.~95, pp. 15--19.
	
	\bibitem{Fan2018}
	Rui Fan, Xiao Ai, and Naim Dahnoun,
	\newblock ``Road surface 3d reconstruction based on dense subpixel disparity
	map estimation,''
	\newblock {\em IEEE Transactions on Image Processing}, vol. 27, no. 6, pp.
	3025--3035, 2018.
	
	\bibitem{Hu2005}
	Zhencheng Hu, Francisco Lamosa, and Keiichi Uchimura,
	\newblock ``A complete uv-disparity study for stereovision based 3d driving
	environment analysis,''
	\newblock in {\em 3-D Digital Imaging and Modeling, 2005. 3DIM 2005. Fifth
		International Conference on}. IEEE, 2005, pp. 204--211.
	
	\bibitem{Ballard1981}
	Dana~H Ballard,
	\newblock ``Generalizing the hough transform to detect arbitrary shapes,''
	\newblock {\em Pattern recognition}, vol. 13, no. 2, pp. 111--122, 1981.
	
	\bibitem{RafaelGonzalez2002}
	C~Rafael~Gonzalez and Richard Woods,
	\newblock ``Digital image processing,''
	\newblock {\em Pearson Education}, 2002.
	
	\bibitem{He2013}
	Kaiming He, Jian Sun, and Xiaoou Tang,
	\newblock ``Guided image filtering,''
	\newblock {\em IEEE transactions on pattern analysis and machine intelligence},
	vol. 35, no. 6, pp. 1397--1409, 2013.
	
	\bibitem{NVIDIA2017}
	NVIDIA,
	\newblock ``Cuda c programming guide,''
	\newblock September 2017.
	
	\bibitem{Andreas2012}
	A~Andreas, Philip Lenz, and Raquel Urtasun,
	\newblock ``Are we ready for autonomous driving? the kitti vision benchmark
	suite,''
	\newblock in {\em Proceedings of the IEEE Conference on Computer Vision and
		Pattern Recognition}, 2012.
	
	\bibitem{Hillel2014}
	Aharon~Bar Hillel, Ronen Lerner, Dan Levi, and Guy Raz,
	\newblock ``Recent progress in road and lane detection: a survey,''
	\newblock {\em Machine vision and applications}, vol. 25, no. 3, pp. 727--745,
	2014.
	
	\bibitem{Geiger2013}
	Andreas Geiger, Philip Lenz, Christoph Stiller, and Raquel Urtasun,
	\newblock ``Vision meets robotics: The kitti dataset,''
	\newblock {\em The International Journal of Robotics Research}, vol. 32, no.
	11, pp. 1231--1237, 2013.
	
\end{thebibliography}

\end{document}